\documentclass[letterpaper]{article} 
\usepackage[preprint]{rffpreprint} 
\usepackage[hyphens]{url}  \usepackage{graphicx}    \usepackage{natbib}  \usepackage{caption} 
\usepackage{algorithm}
\usepackage{algorithmic}

\usepackage{amsmath,amssymb,mathtools}
\DeclareMathSizes{6.4}{6.4}{5}{5}
\DeclareMathSizes{5.8}{5.8}{5}{5}
\DeclareMathSizes{5.2}{5.2}{5}{5}
\usepackage{bm}
\usepackage{subcaption}
\usepackage{booktabs}
\usepackage{multirow}
\usepackage{siunitx}
\usepackage{enumitem}
\usepackage{array}
\usepackage{placeins}

\DeclareMathOperator{\diag}{diag}

\newcommand{\appref}[1]{Appendix \ref{#1}}
\newcommand{\A}{\mathbf{A}}

\newcommand{\bigO}{\mathcal{O}}

\newcommand{\bomega}{\boldsymbol{\omega}}

\newcommand{\GP}{\mathcal{GP}}

\newcommand{\N}{\mathcal{N}}
\newcommand{\eye}{\mathbf{I}}

\newcommand{\E}{\mathbb{E}}

\newcommand{\f}{\mathbf{f}}

\newcommand{\K}{\mathbf{K}}
\newcommand{\bk}{\mathbf{k}}

\newcommand{\calK}{\bm{\mathcal{K}}}

\newcommand{\m}{\mathbf{m}}

\newcommand{\Q}{\mathbf{Q}}
\newcommand{\bq}{\mathbf{q}}
\newcommand{\R}{\mathbb{R}}

\renewcommand{\S}{\mathbf{S}}

\newcommand{\V}{\mathbf{V}}

\newcommand{\W}{\mathbf{W}}
\newcommand{\x}{\mathbf{x}}

\newcommand{\X}{\mathbf{X}}
\newcommand{\y}{\mathbf{y}}
\newcommand{\Y}{\mathbf{Y}}

\newcommand{\inv}{^{-1}}

\title{Random Feature Gaussian Process Attention: \\ Linear-Time Probabilistic Attention with Calibrated Uncertainty}

\author{
Amir Mohammad Mahfoozi\textsuperscript{1\rlap{\smash{,*}}}\quad
Zi Yang\textsuperscript{2\rlap{\smash{,*}}}\quad
Ying Li\textsuperscript{2}\quad
Michael Minyi Zhang\textsuperscript{2}
}
\affiliations{
\textsuperscript{1}Department of Computer Engineering, Sharif University of Technology\\
\textsuperscript{2}School of Computing and Data Science, The University of Hong Kong\\
\texttt{amir.mahfoozi01@sharif.edu \quad \{ziyang2023, lynnli98\}@connect.hku.hk \quad mzhang18@hku.hk}
}

\begin{document}
\pagestyle{plain}
\thispagestyle{plain}

\maketitle
\AddToHookNext{shipout/foreground}{%
  \put(54bp,-716bp){\rule{60bp}{0.4pt}}%
  \put(54bp,-725bp){\normalfont\footnotesize\textsuperscript{*}Equal contribution.}%
}

\begin{abstract}
Transformers provide a state-of-the-art modeling framework, yet poor calibration limits their reliability in safety-critical applications. A promising direction addresses this issue by interpreting attention as a Gaussian process (GP) posterior, which enables principled uncertainty calibration but incurs cubic complexity in sequence length due to the inversion of the kernel; although decoupled GP variants reduced the cost to quadratic, the computation remains prohibitive in practice. In this paper, we propose the plug-and-play random Fourier feature Gaussian process attention (RFF-GPA) module, which represents the attention as a GP with a stationary kernel approximated by random Fourier features. This low-rank approximation results in linear-time complexity for approximating the posterior mean and variance, making it far more scalable compared to previous work. Empirical results on multiple real-world datasets show that our attention module improves calibration while maintaining predictive accuracy, and simultaneously reduces computational complexity to linear in the sequence length.
\end{abstract}

\section{Introduction}\label{sec:intro}
Large language models (LLMs) such as GPT-3 and its successors \citep{brown2020language, zhao2023survey} are built upon the Transformer architecture \citep{vaswani2017attention}. The core of the Transformer is the self-attention mechanism, which enables scalable modeling of long-range dependencies. Transformers have achieved breakthroughs by leveraging this design, not only in natural language processing but also in vision and multimodal learning \citep{vaswani2017attention,dosovitskiy2020image,brown2020language}, thereby becoming the leading architecture in modern artificial intelligence. 

Despite their apparent successes, Transformers are often overconfident in their predictions and lack principled mechanisms for uncertainty quantification, which makes them unreliable in applications where the downstream tasks are consequential \citep{guo2017calibration,minderer2021revisiting, papamarkou2024position}. Recent work has 
bridged the standard attention output to the mean of a Gaussian process (GP) posterior \citep{williams2006gaussian}. Specifically, \citet{chen2023calibrating,bui2025revisiting} describe an exact and a sparse GP formulation where the number of inducing points equals the sequence length $L$. From this perspective, we can extend the framework to incorporate the posterior covariance, which naturally provides a principled and well-calibrated quantification of uncertainty that is absent in vanilla Transformers. 

Nevertheless, existing approaches remain computationally expensive. Typically, inference requires inversion of a $L \times L$-dimensional kernel matrix, resulting in cubic $\bigO(L^3)$ time and quadratic memory complexity, which makes these methods impractical for long sequences in real-world applications. While \citet{chen2023calibrating} introduced a decoupled GP variant to reduce the complexity from cubic to quadratic by decoupling the inducing points for mean and covariance representations, the resulting $\bigO(L^2)$ cost remains impractical for long sequences.

To obtain a model that combines calibrated uncertainty, strong predictive accuracy, and low computational cost, we propose random Fourier feature Gaussian process attention (RFF-GPA), a plug-and-play attention module for Transformer architectures.

\begin{figure*}[t]
    \centering
    \includegraphics[width=\textwidth]{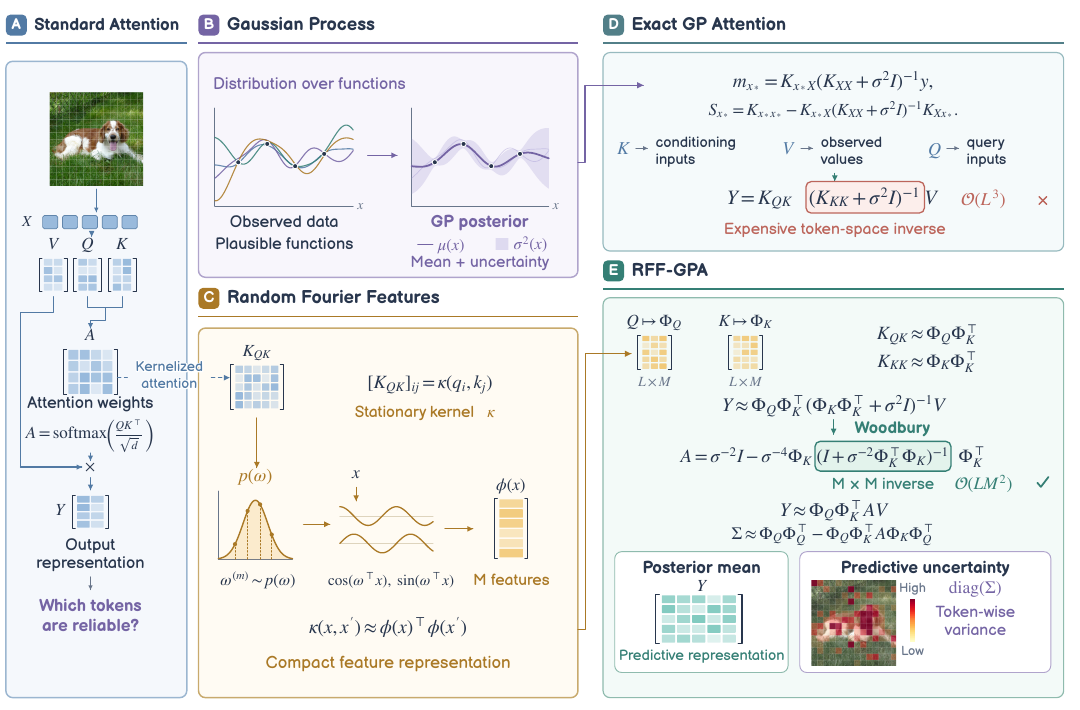}
    \caption{\textbf{Overview of RFF-GPA.}
    (A) Standard attention produces deterministic token representations
    without explicit uncertainty. (B,D) A GP formulation provides
    posterior mean and variance, but exact GP attention requires an
    expensive token-space inverse. (C) Random Fourier features provide
    a compact approximation of the stationary kernel. (E) RFF-GPA
    performs GP inference in the resulting feature space, replacing the
    $L\times L$ inverse with an $M\times M$ inverse while retaining the
    posterior mean and token-wise predictive variance.}
    \label{fig:rff_gpa_overview}
\end{figure*}

Figure~\ref{fig:rff_gpa_overview} summarizes the progression from
standard attention to GP attention and illustrates how our
random-feature formulation preserves posterior uncertainty while
moving the expensive computation from token space to feature space. Our main contributions are summarized as follows:
\begin{itemize}[leftmargin=*]
    \item We introduce RFF-GPA, which approximates the attention kernel with random Fourier features (RFF) \citep{rahimi2007random} and performs GP posterior inference in the resulting feature space. Using the Woodbury matrix identity \citep{hager1989updating}, RFF-GPA avoids direct inversion of the full $L \times L$ kernel matrix and computes predictive means and covariance estimates in linear time with respect to the sequence length.

    \item To relax the symmetry constraint between key and query induced by stationary kernels, we further extend the model to a correlated random-feature GP attention variant, RFF-CGP. This variant models query and key representations through a shared latent GP structure, enabling a more flexible attention mechanism while retaining the computational advantages of the random-feature formulation.

    \item We evaluate the proposed models on image and text classification benchmarks, as well as under distribution shift on CIFAR-10-C. The results show that our methods improve calibration while maintaining competitive accuracy. They also achieve better runtime and memory scaling than inducing-point GP attention, highlighting the favorable accuracy--calibration--efficiency trade-off of our approach.
\end{itemize}

\section{Background}
\label{sec:background}

In this section, we first review the self-attention mechanism in Transformers
and its interpretation as a kernel method 
followed by a brief overview of Gaussian processes and random Fourier features as a scalable kernel approximation.
Together, these components motivate our probabilistic treatment of attention within a Gaussian process framework.

\subsection{Self-Attention}
\label{sec:self-attn}

Given an input sequence $\X=[\x_1,\dots,\x_L]\in\R^{L\times D}$ of $L$ tokens, each represented by a $D$-dimensional vector, the self-attention mechanism first projects the inputs into three different latent spaces~\citep{vaswani2017attention}:
\begin{equation}\label{eq:attn}
	\Q = \X\W_Q, \qquad 
	\K = \X\W_K, \qquad 
	\V = \X\W_V,
\end{equation}
where $\W_Q\in\R^{D\times D_Q}$, $\W_K\in\R^{D\times D_K}$, and $\W_V\in\R^{D\times D_V}$ are learnable weight matrices that project the input $\X$ into $\Q$, $\K$, and $\V$, referred to as queries, keys, and values, respectively. In standard self-attention, we set the latent dimensionalities of the queries, keys and values to be equal, $D_Q=D_K=D_V=D$, and let $\W_Q=\W_K$, yielding $\mathbf{Q}=\mathbf{K}$ for simplicity~\citep{vaswani2017attention}.

Then, the attention matrix is
\begin{equation}
	\A=\operatorname{softmax}\!\big(\Q\K^\top/\sqrt{D_K}\big),
\end{equation}
where the softmax is applied row-wise. The output is $\mathbf{Y}=\mathbf{A}\mathbf{V}$. Here, each row of $\mathbf{A}$ specifies how a token (i.e., an element of the input sequence) aggregates information from all others, weighted by their query–key similarity.

Furthermore, multi-head self-attention (MHSA) applies the above operation in parallel across $H$ heads, each with its own parameter set $(\W_Q^{(h)}, \W_K^{(h)}, \W_V^{(h)})$. For the $h$-th head, the output is
\begin{equation}
	\Y^{(h)} = \operatorname{softmax}\!\big(\Q^{(h)}\K^{(h)\top}/\sqrt{D_K}\big)\,\V^{(h)} ,
\end{equation}
and the final output is obtained by concatenating all head outputs and projecting back:
\begin{equation}
	\mathbf{H} = \operatorname{concat}\big(\mathbf{Y}^{(1)}, \dots, \mathbf{Y}^{(H)}\big)\mathbf{W}_o,
\end{equation}
where $\mathbf{W_o}\in\mathbb{R}^{(HD_V)\times D}$ is a learnable projection. This design allows different heads to capture diverse similarity patterns among tokens, enhancing the expressiveness of the model.

\subsection{Kernel Attention}
\label{sec:kernel-attn}

The attention matrix can be further viewed as a softmax similarity with a positive-definite kernel, $\kappa(\cdot,\cdot)$. \citet{tsai2019transformer} established a direct connection between the attention module and the classic kernel methods. 
Generally, we may represent the attention module as a kernel function:
\begin{align}
	\calK_{QK}&\in\R^{L\times L}, \nonumber \\ 
	[\calK_{QK}]_{ij}&=\kappa(\bq_i,\bk_j)=\kappa(\x_i \W_{Q} , \x_j \W_{K}),
\end{align}
and the outputs of the Transformer take the form
\begin{equation}
    \label{eq:kernel_attention}
	\Y =  \calK_{QK} \,\V. 
\end{equation}
Note that, to guarantee kernel symmetry in the attention mechanism, we must
enforce
\begin{align}
   \kappa(\x_i \W_{Q}, \x_j \W_{K}) \;=\; \kappa(\x_j \W_{K}, \x_i \W_{Q}),
\end{align}
that requires $\W_Q = \W_K$, which in turn reduces the expressive capacity of the Transformer.  
Moreover, the connection (Eq.~\ref{eq:kernel_attention}) between kernels and attention modules is crucial for making the probabilistic connection between Transformers and GPs.

\subsection{Gaussian Processes}\label{sec:gp}

A Gaussian process (GP), $$f \sim \GP\left(0,\kappa(\cdot,\cdot)\right),$$ is a distribution over functions $f:\R^D \to \R$ such that any finite collection of function values is jointly Gaussian~\citep{williams2006gaussian}. Formally, for inputs $\X=\left[ \x_1, \ldots , \x_N \right]$, we have
\begin{equation}
\f|\X \sim \N\left(0,\calK_{XX}\right).
\end{equation} 
We assume we have noisy observations $y_i=f(\x_i)+\varepsilon_i$ with $\varepsilon_i \sim \N(0,\sigma^2)$. We let $\y=[y_1,\dots,y_N]^\top$ and we define the Gram matrix, $\calK_{XX}\in\R^{N\times N}$, to have entries $[\calK_{XX}]_{ij}=\kappa(\x_i,\x_j)$. For test inputs $\X_\star$, the predictive posterior distribution is a multivariate Gaussian distribution, $\N(\m_{X_\star}, \S_{X_\star})$,  with mean and variance:
\begin{align}
	\m_{X_\star}
	&= \calK_{X_\star X}\big(\calK_{XX}+\sigma^2 \eye \big)\inv\y, \label{eq:gp_mean} \\
	\S_{X_\star}
	&= \calK_{X_\star X_\star}
	- \calK_{X_\star X}\big(\calK_{XX}+\sigma^2 \eye\big)\inv\calK_{XX_\star}.
\end{align}

\subsection{Random Fourier Features}
\label{sec:rff}

Random Fourier features are a technique to accelerate computation of kernel methods through a low-rank factorization. The basic theorem necessary to approximate a stationary kernel is Bochner’s theorem~\citep{bochner1959lectures}, which states that any continuous shift-invariant kernel, $\kappa(\x,\x') = \kappa(\x-\x')$, can be expressed as the Fourier transform of a nonnegative spectral density $p(\bomega)$, namely:
\begin{equation}
\kappa(\x-\x') = \int p(\bm \omega)\, e^{i \bm \omega^\top (\x-\x')} \,\mathrm{d}\bm \omega. 
\end{equation}
Based on this result, \citet{rahimi2007random} proposed to first draw $M/2$ frequencies $\bm \Omega \triangleq \{\bomega^{(m)}\}_{m=1}^{M/2}\!\sim p(\bomega)$, and then obtain an unbiased estimator for kernel function: 
\begin{equation}
\kappa(\x,\x') \;=\; \E\left[\phi(\x)^\top \phi(\x')\right],
\label{eq:rff-approx}
\end{equation} 
where the randomized feature mapping is:
\begin{equation}
\phi(\x) = \sqrt{\tfrac{2}{M}}
\begin{bmatrix}
\cos(\bm \omega^{(1)\top} \x) \\
\sin(\bm \omega^{(1)\top}\x) \\
\vdots \\
\cos(\bm \omega^{(M/2)\top}\x) \\
\sin(\bm \omega^{(M/2)\top}\x)
\end{bmatrix}.
\label{eq:basis_fun}
\end{equation}
As a result, the full kernel matrix $\calK_{\X\X}$ is approximated by $\Phi(\X)\Phi(\X)^\top$ with $\Phi(\X)=\left[\phi(\x_1),\ldots,\phi(\x_N)\right]^\top$, allowing scalable GP computations through low-rank approximations instead of the direct inversion of $\calK_{\X\X}$.

\subsection{Related Works}\label{sec:related_works}
Subsequent extensions of the Transformer have accordingly combined the RFF approximations and the attention kernel. \citet{pengrandom} introduced random feature attention (RFA) as a direct replacement of the linear projections of the queries and keys in the original Transformer with the basis function in Eq.~\eqref{eq:basis_fun}. In parallel, \citet{choromanski2020rethinking} introduced positivity and orthogonality constraints for the random feature approximation of the attention module. These approaches are efficient and but they only model the interaction between the queries and keys as a kernel. Moreover, they do not pose the Transformer as a fully Bayesian model and therefore lack principled uncertainty quantification.

In Bayesian deep learning, \citet{neal1995bayesian} originally established the connection between Gaussian processes and neural networks. \citet{damianou2013deep} then introduced the deep Gaussian process, which applies the feedforward network structure to Gaussian process regression. \citet{gal2016dropout} further established the connection between training deep neural networks with approximate Bayesian inference for deep Gaussian processes.

Comparatively, the Bayesian perspective on Transformers remains underexplored. 
\citet{tran2019bayesian,fan2020bayesian,xue2021bayesian} implemented Bayesian Transformers with varying architectures and priors on the network parameters, but these methods overlook the kernelized attention structure which limits their modeling capacity. As we will demonstrate, unifying attention modules, random feature kernel approximations, and Gaussian processes will be crucial to obtain properly calibrated probabilistic Transformers.

\section{Methodology}
\label{sec:method}

This section first presents the proposed random Fourier feature Gaussian process attention module (Sec.~\ref{sec:rffgpa}), then details its integration into the standard Transformer (Sec.~\ref{sec:training_objective}), and finally provides an illustrative implementation in GP-based Transformers (Sec.~\ref{sec:rff_cgp}).

\subsection{Random Fourier Feature Gaussian Process Attention (RFF-GPA)} 
\label{sec:rffgpa}

In a typical GP-based Transformer \citep{bui2025revisiting, chen2023calibrating}, the output of the kernel attention module is modeled as the predictive posterior mean of a Gaussian process. We interpret the keys as the ``training inputs'', the value matrix $\V = \X \W_{V}^{\top}$ as the corresponding (noisy) observations, and the queries as the ``test inputs''. Let \(\calK_{KK}\) denote the kernel Gram matrix with keys as inputs. Denoting the \(d\)-th column of \(\Y\) and \(\V\) as \(\y_d\) and \(\mathbf{v}_d\), respectively, the \(d\)-th column of the kernel attention output is precisely the GP predictive posterior mean (see Eq.~\ref{eq:gp_mean}):
\begin{align}
    \label{eq:attention_mean}
    \y_d = \calK_{QK} \left( \calK_{KK} + \sigma^2 \mathbf{I} \right)\inv \mathbf{v}_d.
\end{align}
Based on this interpretation, it is straightforward to obtain the attention covariance as
\begin{align}
    \label{eq:attention_cov}
    \bm \Sigma = \calK_{QQ} - \calK_{QK} \left( \calK_{KK} + \sigma^2 \mathbf{I} \right)^{-1} \calK_{KQ}.
\end{align}
The overall output of the kernel attention module~$\Y$ can thus be interpreted as the predictive mean of $D$ independent GPs, with the corresponding predictive covariance capturing the associated uncertainty. This view extends naturally to multi-head self-attention, where the predictive outputs from all heads are concatenated and mapped through the output projection~$\W_{o}$.

Unfortunately, inverting the kernel Gram matrix $\calK_{KK}$ incurs cubic complexity, which is impractical for long sequences. Two common strategies to reduce the cost of GP inference are: (i) inducing points (sparse GPs) and (ii) RFFs. \citet{chen2023calibrating} interpret the attention module as a sparse GP by treating queries as inducing points, which makes the number of inducing points equal to $L$ but still results in $\mathcal{O}(L^3)$ complexity. 
\begin{algorithm}[t]
\caption{RFF-GPA (Single-Head): Random Fourier Feature Gaussian Process Attention}
\label{alg:rffgpa}
\begin{algorithmic}[1]
\REQUIRE Sequence of tokens $\X = (\x_1,\ldots,\x_L)$
\STATE Compute $\Q, \K, \V$ using Eq.~\eqref{eq:attn}.
\STATE Draw $S$ samples $\hat{\bm{\Omega}}^{(s)}$ from $p(\bm{\Omega})$ using the reparameterization trick.
\STATE Compute the feature maps $\Phi_Q$ and $\Phi_K$ using $\Q$, $\K$, and sampled spectral points $\hat{\bm{\Omega}}^{(s)}$.
\STATE Compute $\Y$ and $\bm{\Sigma}$ using Eq.~\eqref{eq:final_cov}.
\STATE Draw $S$ samples $\hat{\Y}^{(s)}$ from $p(\Y \mid \bm{\Omega}, \X)$ using the reparameterization trick.
\STATE Evaluate the ELBO using Eq.~\eqref{eq:elbo}.
\STATE Maximize the ELBO and update $\bm{\Theta}, \bm{\zeta}$ using Adam \citep{kingma2015adam}.
\ENSURE Model parameters $\bm{\Theta}$ and variational parameters $\bm{\zeta}$.
\end{algorithmic}
\end{algorithm}
Alternatively, we turn to RFFs (Eq.~\ref{eq:rff-approx}), where the kernel Gram matrices can be approximated as \(\calK_{KK} \approx \Phi_K \Phi_K^\top\) and \(\calK_{QK}\approx \Phi_Q  \Phi_K^\top\). 
Substituting those approximations into Eq.~\eqref{eq:attention_mean} and Eq.~\eqref{eq:attention_cov} yields\footnote{For clarity, the following derivations focus on a single head, but the extension to the multi-head case is straightforward.}: 
\begin{align}
\label{eq:appro_mean_cov}
\Y &\approx\; \Phi_Q \Phi_K^\top \,\big(\Phi_K \Phi_K^\top + \sigma^2 \mathbf{I}\big)^{-1} \V, \\ 
\mathcal{V}(\y_d) &\approx \Phi_Q \Phi_Q^\top \!-\! \Phi_Q \Phi_K^\top \big(\Phi_K \Phi_K^\top \!+\! \sigma^2 \mathbf{I}\big)^{-1} \Phi_K \Phi_Q^\top. \nonumber  
\end{align}
For simplicity, we denote the computational demanding term as 
\begin{equation}
    \A \triangleq \left( \Phi_K \Phi_K^\top + \sigma^2 \mathbf{I} \right)^{-1}.  
\end{equation}
By exploiting the low-rank structure of $\Phi_K$, we can apply the Woodbury identity and obtain 
\begin{align}
    \label{eq:woodbury}
    \A &= \tfrac{1}{\sigma^2}\mathbf{I} 
    - \tfrac{1}{\sigma^4}\Phi_K 
    \Big( \mathbf{I} + \tfrac{1}{\sigma^2}\Phi_K^\top \Phi_K \Big)^{-1} \Phi_K^\top, 
\end{align} 
where we only need to invert an $M \times M$ matrix, with $M$ pre-selected, thereby reducing the overall cost to $\mathcal{O}(LM^2)$, i.e., linear in $L$. 

Replacing the Eq.~\eqref{eq:woodbury} within Eq.~\eqref{eq:appro_mean_cov}, we can obtain
\begin{align}
    \Y &\approx \Phi_Q \Phi_K^\top \A  \V, \label{eq:final_mean} \\ 
    \bm \Sigma &\approx \Phi_Q \Phi_Q^\top \!-\! \Phi_Q \Phi_K^\top \A \Phi_K \Phi_Q^\top. \label{eq:final_cov}
\end{align}
Instead of predicting the full covariance function, we only report the diagonal variance, denoted as $\operatorname{diag}(\bm \Sigma)$, that represents the uncertainty for each token.

In supervised Transformer applications, the latent Gaussian attention output
$\Y$ serves as the representation passed to the final prediction head.
Consequently, the observed labels are modeled through a likelihood
$p(\bm{\mathcal{Y}} \mid \Y)$, yielding the overall supervised joint distribution
\begin{equation}
p(\bm{\mathcal{Y}}, \Y, \bm\Omega \mid \X)
=
p(\bm{\mathcal{Y}} \mid \Y)\,
p(\Y \mid \X, \bm\Omega)\,
p(\bm\Omega),
\end{equation}
where $p(\bm{\mathcal{Y}} \mid \Y)$ denotes the categorical likelihood induced
by the prediction head.

\subsection{Variational Inference for Standard Transformers with RFF-GPA Module} 
\label{sec:training_objective}

The RFF-GPA module introduces latent attention outputs and spectral points whose exact posterior is not available in closed form, so we use variational inference to obtain a tractable training objective.
We now describe how to train a Transformer equipped with the proposed RFF-GPA
module within a variational inference (VI) framework. Our goal is to
jointly learn the model parameters\footnote{Note that, for notational simplicity, $\W_Q, \W_K, \W_V, \W_o$ now denote the
corresponding parameters for all layers and heads.}, 
$\bm \Theta = \{\W_Q, \W_K, \W_V, \W_o, \sigma^2\}$,
and infer the posterior distribution of the spectral points $\bm\Omega$ given the observations $\bm{\mathcal{Y}}$. Since the exact posterior $p(\bm \Omega \mid \bm{\mathcal{Y}})$ is intractable, VI posits a tractable family $q(\bm \Omega)$ to approximate it and optimizes both the variational parameters of $q(\bm \Omega)$ and the model parameters $\bm \Theta$ by maximizing the evidence lower
bound (ELBO). 

Following the settings in~\citet{li2024preventing, yang2025multi}, we assume that
the variational distribution $q(\bm \Omega)$ coincides with the prior\footnote{See Appendix~\ref{app:empirical_bayes} for an empirical Bayes view.},
\begin{equation}
p(\bm \Omega) \triangleq \prod_{m=1}^{M} \mathcal{N}(\bm \mu, \mathbf{S}),
\end{equation}
where $\bm \mu$ denotes the mean and $\mathbf{S}$ is a diagonal covariance matrix.
Consequently, the ELBO is: 
\begin{align}
\mathcal{L}_{\mathrm{ELBO}}
&\!=\!
\mathbb{E}_{p(\bm \Omega)}\,
\mathbb{E}_{\,p(\mathbf{Y}\mid \bm \Omega, \X)}\!
\left[\log p\!\left(\bm{\mathcal{Y}} \mid \Y, \X, \bm \Omega \right)\right], 
\end{align} 
where the expectation over the log-likelihood, $\log p(\mathbf{Y}\mid \bm \Omega, \X)$, is required since the attention output $\mathbf{Y}$ is a random latent variable rather than a deterministic value, with its mean and covariance given by Eq.~\eqref{eq:final_mean} and Eq.~\eqref{eq:final_cov}, respectively. 

This term indeed measures the expected reconstruction accuracy. For
evaluation, we employ the reparameterization trick to draw $S$ samples from
$p(\bm \Omega)$ and $p(\mathbf{Y}\mid \bm \Omega, \X)$:
\begin{align}
\bm \omega^{(s)}_m &= \bm \mu + \sqrt{\mathbf{S}} \bm \xi^{(s)}, 
& \bm \xi^{(s)} \sim \mathcal{N}(\mathbf{0}, \mathbf{I}), \\
\hat{\y}^{(s)}_d &= \mathbf{y}_d + \sqrt{\operatorname{diag}(\bm \Sigma)} \odot \bm{\varepsilon}^{(s)}, 
& \bm{\varepsilon}^{(s)} \sim \mathcal{N}(\mathbf{0}, \mathbf{I}),
\end{align} 
and thereby obtain an unbiased Monte Carlo estimator of the reconstruction
term:
\begin{equation}\label{eq:elbo}
\mathcal{L}_{\mathrm{ELBO}} \approx \sum_{s=1}^{S}
\left[\log p\!\left(\bm{\mathcal{Y}} \mid \hat{\Y}^{(s)}, \X, \hat{\bm \Omega}^{(s)} \right)\right], 
\end{equation}
Consequently, we optimize the above objective with respect to both the model
parameters $\bm \Theta = \{\W_Q, \W_K, \W_V, \W_o, \sigma^2\}$ and the
variational parameters $\bm \zeta \triangleq \{ \bm \mu, \mathbf{S} \}$. 
The complete procedure is summarized in
Algorithm~\ref{alg:rffgpa}. 
\subsection{Random Fourier Feature Correlated Gaussian Process Attention (RFF-CGP)} 
\label{sec:rff_cgp}
Although RFF-GPA provides scalable uncertainty-aware attention, its stationary-kernel construction still inherits a symmetry constraint from standard kernel attention: the query and key projections are tied through the same kernel structure, limiting the ability to model asymmetric query-key interactions. This can be undesirable because queries and keys play different roles in attention, and forcing them into a symmetric kernel space can restrict representation capacity. To address this limitation, we extend our random-feature formulation to correlated Gaussian process attention (CGP) \citep{bui2025revisiting}, yielding random Fourier feature correlated Gaussian process attention (RFF-CGP).

CGP attention models the self-attention output as the cross-covariance between 
two correlated Gaussian processes, each defined by an affine transformation of 
a latent canonical GP. This construction naturally removes the requirement that 
$\W_Q=\W_K$, which allows asymmetric kernels and enhances representation 
capacity. In particular, the $d$-th output column can be expressed as
\begin{equation}
    \y_d\!=\!\K_{QO}\,(\K_{OO}+\sigma^2 I)^{-1}\,
           \K_{OK}\,(\K_{KK}+\sigma^2 I)^{-1}\,\mathbf{v}_d,
    \label{eq:cgp-mean}
\end{equation}
where $\K_{QO}, \K_{OK}, \K_{KK}, \K_{OO}$ are the Gram or cross-covariance matrices
induced by the canonical kernel, and $\mathbf{v}_d$ is the $d$-th column of the value matrix $\V$, as in
Eq.~\eqref{eq:attention_mean}.

To make Eq.~\eqref{eq:cgp-mean} scalable, we first substitute each kernel matrix with 
its RFF approximations. Let $\Phi_Q, \Phi_K, \Phi_O \in \mathbb{R}^{L \times M}$ 
be the RFF embeddings of queries, keys, and the latent canonical GP inputs, 
respectively. Then, we have: 
\begin{align}
    \begin{aligned}
    \K_{QO} &\approx \Phi_Q \Phi_O^\top, &
    \K_{OK} &\approx \Phi_O \Phi_K^\top, \\
    \K_{KK}  &\approx \Phi_K \Phi_K^\top, &
    \K_{OO}  &\approx \Phi_O \Phi_O^\top .
    \end{aligned}
\end{align}
Substituting into Eq.~\eqref{eq:cgp-mean} gives the following predictive mean: 
\begin{small}
    \begin{equation}
    \y_d \approx\Phi_Q \Phi_O^\top 
    \left(\Phi_O \Phi_O^\top\!+\!\sigma^2 I\right)\!^{-1}
    \Phi_O \Phi_K^\top 
    \left(\Phi_K \Phi_K^\top\!+\!\sigma^2 I\right)\!^{-1}
    \mathbf{v}_d.
    \label{eq:rff-cgp-mean}
    \end{equation} 
\end{small}

Both inverses in Eq.~\eqref{eq:rff-cgp-mean} involve matrices of size $L \times L$. 
Using the Woodbury identity, they can be rewritten as inversions of 
$M \times M$ matrices:
\begin{small}
    \begin{align}
    \big(\Phi_K \Phi_K^\top+ \sigma^2 I\big)^{-1}
    &= \tfrac{1}{\sigma^2}I 
     - \tfrac{1}{\sigma^4}\Phi_K \Big(I+\tfrac{1}{\sigma^2}\Phi_K^\top \Phi_K\Big)^{-1}\Phi_K^\top, \nonumber \\
    \big(\Phi_O \Phi_O^\top + \sigma^2 I\big)^{-1}
    &= \tfrac{1}{\sigma^2}I 
     - \tfrac{1}{\sigma^4}\Phi_O \Big(I+\tfrac{1}{\sigma^2}\Phi_O^\top \Phi_O\Big)^{-1}\Phi_O^\top. \nonumber
    \end{align} 
\end{small}
Analogous to Eq.~\eqref{eq:final_cov}, the predictive covariance of RFF-CGP can be derived in closed form by substituting the RFF approximations into the CGP conditional covariance \cite{bui2025revisiting}. Specifically, in the CGP construction, the query process is conditioned on the canonical latent GP $O$; the resulting token-wise predictive variance is therefore the residual query variance after this conditioning,
\begin{align}
\operatorname{diag}(\bm \Sigma) = \sigma^2 \operatorname{diag}\!\Big( \Phi_Q \big(\Phi_O^\top \Phi_O + \sigma^2 I\big)^{-1} \Phi_Q^\top \Big),
\label{eq:rff-cgp-cov}
\end{align}
which is the exact one-step posterior variance and reduces to the RFF-GPA variance in Eq.~\eqref{eq:final_cov} when $\Phi_O\!\to\!\Phi_Q$ (i.e. $\K_{OO}\!\to\!\K_{QQ}$). The data-dependent terms of the full CGP conditional covariance cancel against the mean outer product, so Eq.~\eqref{eq:rff-cgp-cov} is independent of $\V$, positive semi-definite, and computed in $\mathcal{O}(LM^2)$ via the same $M\times M$ inversion $(\Phi_O^\top\Phi_O+\sigma^2 I)^{-1}$ as in Eq.~\eqref{eq:rff-cgp-mean}.

\section{Experiments}
\label{sec:experiments}
We first evaluate our proposed method\footnote{
Appendix~\ref{app:baselines} provides detailed descriptions of all baseline methods, together with their training configurations and hyperparameter settings.} (RFF-GPA and RFF-CGP) on in-distribution accuracy and calibration in Sec.~\ref{subsec:ID}. We then demonstrate its efficiency and scalability in terms of computation and memory on synthetic sequence benchmarks in Sec.~\ref{subsec:efficiency}, and finally evaluate out-of-distribution robustness under distribution shift on CIFAR-10-C in Sec.~\ref{subsec:OUT}.
\begin{table*}[t!]
\centering
\setlength{\tabcolsep}{1.5mm}
\resizebox{\textwidth}{!}{
\begin{tabular}{c||ccc|ccc}
\toprule
\multirow{2}{*}{METHOD}
& \multicolumn{3}{c|}{IMAGE}
& \multicolumn{3}{c}{TEXT} \\
\cmidrule(lr){2-4} \cmidrule(lr){5-7}
& F-MNIST & CIFAR-10 & SVHN & 20NG & Hyperp. & SST-2 \\
\midrule
\multirow{1}{*}{METRIC} & \multicolumn{6}{||c}{ACC~$\uparrow$} \\ \midrule\midrule
MLE       & 0.889 $\pm$ 0.003 & 0.715 $\pm$ 0.008 & 0.909 $\pm$ 0.007 & 0.654 $\pm$ 0.015 & 0.744 $\pm$ 0.015 & 0.775 $\pm$ 0.009 \\
MLE+Temp  & 0.889 $\pm$ 0.003 & 0.715 $\pm$ 0.008 & 0.909 $\pm$ 0.007 & 0.654 $\pm$ 0.015 & 0.744 $\pm$ 0.015 & 0.775 $\pm$ 0.009 \\
MCD       & 0.885 $\pm$ 0.001 & 0.713 $\pm$ 0.002 & 0.902 $\pm$ 0.001 & \textbf{0.682 $\pm$ 0.004} & 0.785 $\pm$ 0.010 & 0.781 $\pm$ 0.006 \\
SNGP      & \underline{0.903 $\pm$ 0.002} & 0.727 $\pm$ 0.007 & \underline{0.913 $\pm$ 0.004} & \underline{0.680 $\pm$ 0.009} & 0.774 $\pm$ 0.015 & \underline{0.789 $\pm$ 0.004} \\
SGPA      & 0.888 $\pm$ 0.002 & 0.665 $\pm$ 0.017 & 0.851 $\pm$ 0.007 & \underline{0.680 $\pm$ 0.002} & 0.749 $\pm$ 0.036 & 0.785 $\pm$ 0.004 \\
RFF-CGP (ours) & 0.902 $\pm$ 0.002 & \underline{0.737 $\pm$ 0.003} & 0.910 $\pm$ 0.006 & 0.651 $\pm$ 0.020 & \textbf{0.815 $\pm$ 0.036} & \textbf{0.790 $\pm$ 0.009} \\
\textbf{RFF-GPA (ours)} & \textbf{0.915 $\pm$ 0.003} & \textbf{0.759 $\pm$ 0.001} & \textbf{0.915 $\pm$ 0.000} & \textbf{0.682 $\pm$ 0.010} & \underline{0.800 $\pm$ 0.015} & \underline{0.789 $\pm$ 0.010} \\
\midrule
\multirow{1}{*}{METRIC} & \multicolumn{6}{||c}{NLL~$\downarrow$} \\ \midrule\midrule
MLE       & 0.317 $\pm$ 0.005 & 0.837 $\pm$ 0.022 & 0.316 $\pm$ 0.020 & 1.483 $\pm$ 0.045 & 0.538 $\pm$ 0.023 & 0.599 $\pm$ 0.042 \\
MLE+Temp  & 0.304 $\pm$ 0.008 & 0.814 $\pm$ 0.024 & 0.305 $\pm$ 0.019 & \textbf{1.224 $\pm$ 0.041} & 0.527 $\pm$ 0.030 & 0.555 $\pm$ 0.017 \\
MCD       & 0.317 $\pm$ 0.004 & 0.819 $\pm$ 0.012 & 0.323 $\pm$ 0.006 & 1.388 $\pm$ 0.025 & 0.558 $\pm$ 0.018 & \underline{0.535 $\pm$ 0.018} \\
SNGP      & \underline{0.282 $\pm$ 0.008} & 0.807 $\pm$ 0.014 & \underline{0.299 $\pm$ 0.007} & 1.381 $\pm$ 0.052 & 0.511 $\pm$ 0.027 & 0.641 $\pm$ 0.045 \\
SGPA      & 0.313 $\pm$ 0.003 & 0.948 $\pm$ 0.043 & 0.483 $\pm$ 0.024 & \underline{1.291 $\pm$ 0.010} & 0.548 $\pm$ 0.046 & 0.592 $\pm$ 0.057 \\
RFF-CGP (ours) & 0.284 $\pm$ 0.003 & \underline{0.802 $\pm$ 0.005} & 0.313 $\pm$ 0.006 & 1.643 $\pm$ 0.173 & \textbf{0.415 $\pm$ 0.037} & \textbf{0.532 $\pm$ 0.043} \\
\textbf{RFF-GPA (ours)} & \textbf{0.252 $\pm$ 0.003} & \textbf{0.713 $\pm$ 0.007} & \textbf{0.292 $\pm$ 0.001} & 1.316 $\pm$ 0.050 & \underline{0.501 $\pm$ 0.007} & 0.546 $\pm$ 0.062 \\
\midrule
\multirow{1}{*}{METRIC} & \multicolumn{6}{||c}{ECE~$\downarrow$} \\ \midrule\midrule
MLE       & 0.051 $\pm$ 0.005 & 0.065 $\pm$ 0.005 & 0.030 $\pm$ 0.003 & 0.158 $\pm$ 0.015 & 0.120 $\pm$ 0.009 & 0.135 $\pm$ 0.017 \\
MLE+Temp  & 0.031 $\pm$ 0.003 & 0.062 $\pm$ 0.001 & 0.027 $\pm$ 0.001 & 0.152 $\pm$ 0.006 & \underline{0.114 $\pm$ 0.001} & 0.127 $\pm$ 0.009 \\
MCD       & 0.030 $\pm$ 0.002 & 0.064 $\pm$ 0.004 & 0.025 $\pm$ 0.002 & 0.136 $\pm$ 0.006 & 0.156 $\pm$ 0.010 & 0.107 $\pm$ 0.006 \\
SNGP      & \underline{0.026 $\pm$ 0.006} & 0.063 $\pm$ 0.006 & 0.017 $\pm$ 0.003 & 0.156 $\pm$ 0.007 & 0.122 $\pm$ 0.035 & 0.132 $\pm$ 0.009 \\
SGPA      & \textbf{0.021 $\pm$ 0.002} & \textbf{0.025 $\pm$ 0.013} & \underline{0.016 $\pm$ 0.005} & \underline{0.134 $\pm$ 0.003} & 0.136 $\pm$ 0.073 & 0.127 $\pm$ 0.016 \\
RFF-CGP (ours) & 0.029 $\pm$ 0.004 & 0.066 $\pm$ 0.013 & 0.023 $\pm$ 0.008 & 0.137 $\pm$ 0.034 & \textbf{0.111 $\pm$ 0.028} & \textbf{0.101 $\pm$ 0.014} \\
\textbf{RFF-GPA (ours)} & \textbf{0.021 $\pm$ 0.003} & \underline{0.046 $\pm$ 0.003} & \textbf{0.013 $\pm$ 0.001} & \textbf{0.130 $\pm$ 0.010} & 0.123 $\pm$ 0.017 & \underline{0.105 $\pm$ 0.011} \\
\bottomrule
\end{tabular}
}
\caption{In-distribution accuracy and calibration on three image and three text classification benchmarks. Each column reports one dataset; row blocks report ACC~$\uparrow$, NLL~$\downarrow$, and ECE~$\downarrow$. Mean and standard deviation are computed over five runs. The best result per column is \textbf{bolded} and the second-best is \underline{underlined}.}
\label{table:id_calibration}
\end{table*}

\subsection{In-Distribution Accuracy and Calibration}\label{subsec:ID}
We first evaluate in-distribution accuracy and calibration across a broad suite of six classification benchmarks spanning both modalities: three image datasets (Fashion-MNIST, CIFAR-10, SVHN) and three text datasets (20~Newsgroups, Hyperpartisan, SST-2).\footnote{Full dataset descriptions, splits, and preprocessing are provided in \appref{app:datasets}.} All methods share an identical Transformer backbone and training protocol, differing only in the attention/output uncertainty mechanism, so that differences in Table~\ref{table:id_calibration} reflect the uncertainty mechanism rather than backbone architecture or tuning. We compare our random-feature GP attention (RFF-GPA and RFF-CGP) against a deterministic maximum-likelihood head \citep[][MLE]{vaswani2017attention}, its post-hoc temperature-scaled variant \citep[][MLE+Temp]{guo2017calibration}, Monte~Carlo dropout \citep[][MCD]{gal2016dropout}, the spectral-normalized neural GP \citep[][SNGP]{liu2020simple}, and sparse GP attention \citep[][SGPA]{chen2023calibrating}. Accuracy is measured by top-1 accuracy (ACC; higher is better), and calibration by negative log–likelihood (NLL) and expected calibration error (ECE), both lower being better.

\paragraph{Accuracy.} RFF-GPA is strongest on the image benchmarks, attaining the best accuracy on Fashion-MNIST ($0.915$), CIFAR-10 ($0.759$), and SVHN ($0.915$), while also matching the best mean accuracy on 20~Newsgroups. On the text benchmarks, the two random-feature variants are complementary: RFF-CGP achieves the best accuracy on Hyperpartisan ($0.815$) and SST-2 ($0.790$), while RFF-GPA remains second-best on both. Notably, RFF-GPA substantially improves over SGPA on the image datasets (e.g.\ $+9.4$ points on CIFAR-10 and $+6.4$ on SVHN), indicating that the random-feature GP formulation not only reduces cost but can also yield a more effective attention mechanism than inducing-point sparse GPs.

\begin{figure*}[t!]
  \centering
  \includegraphics[width=.48\linewidth]{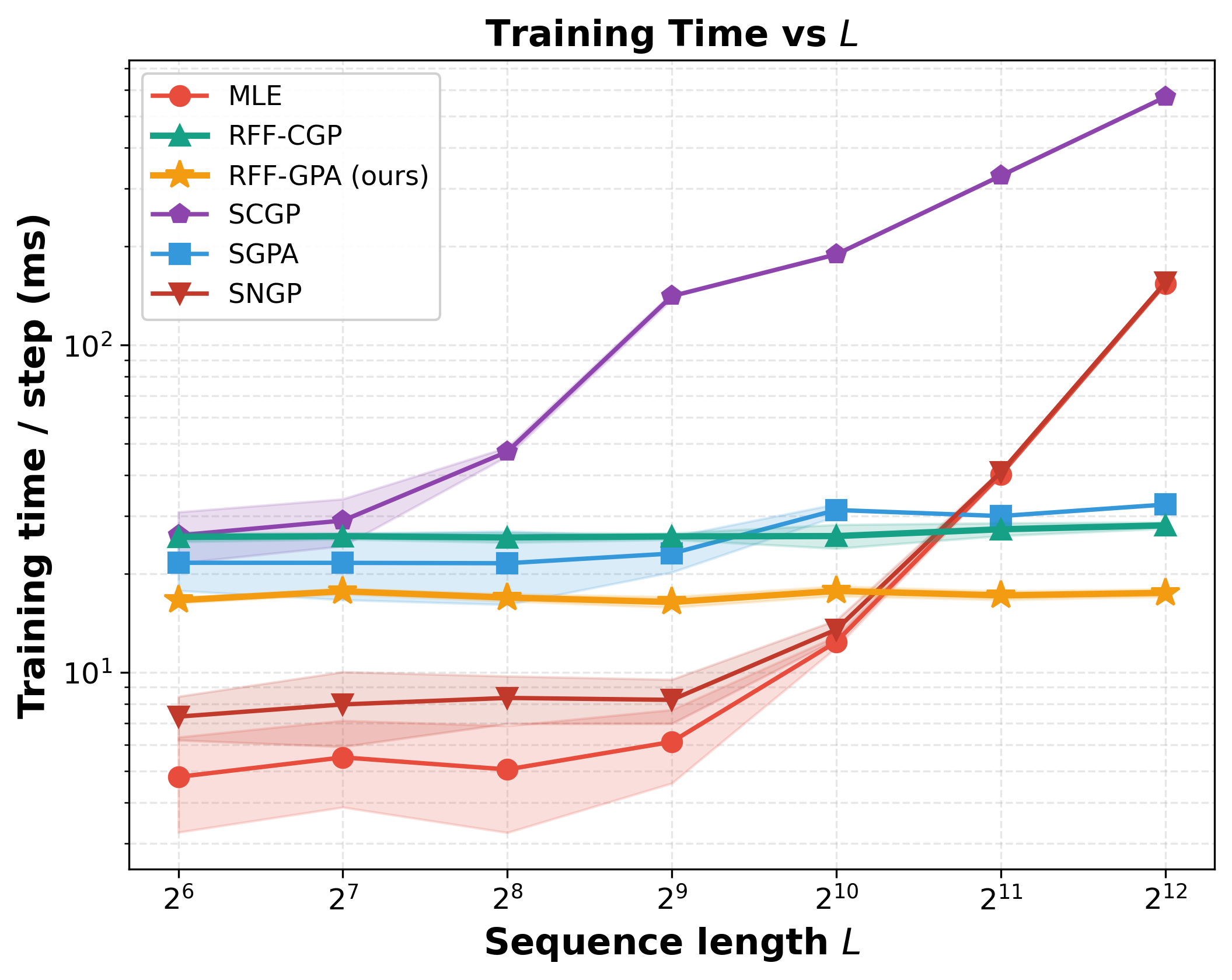}\hfill
  \includegraphics[width=.48\linewidth]{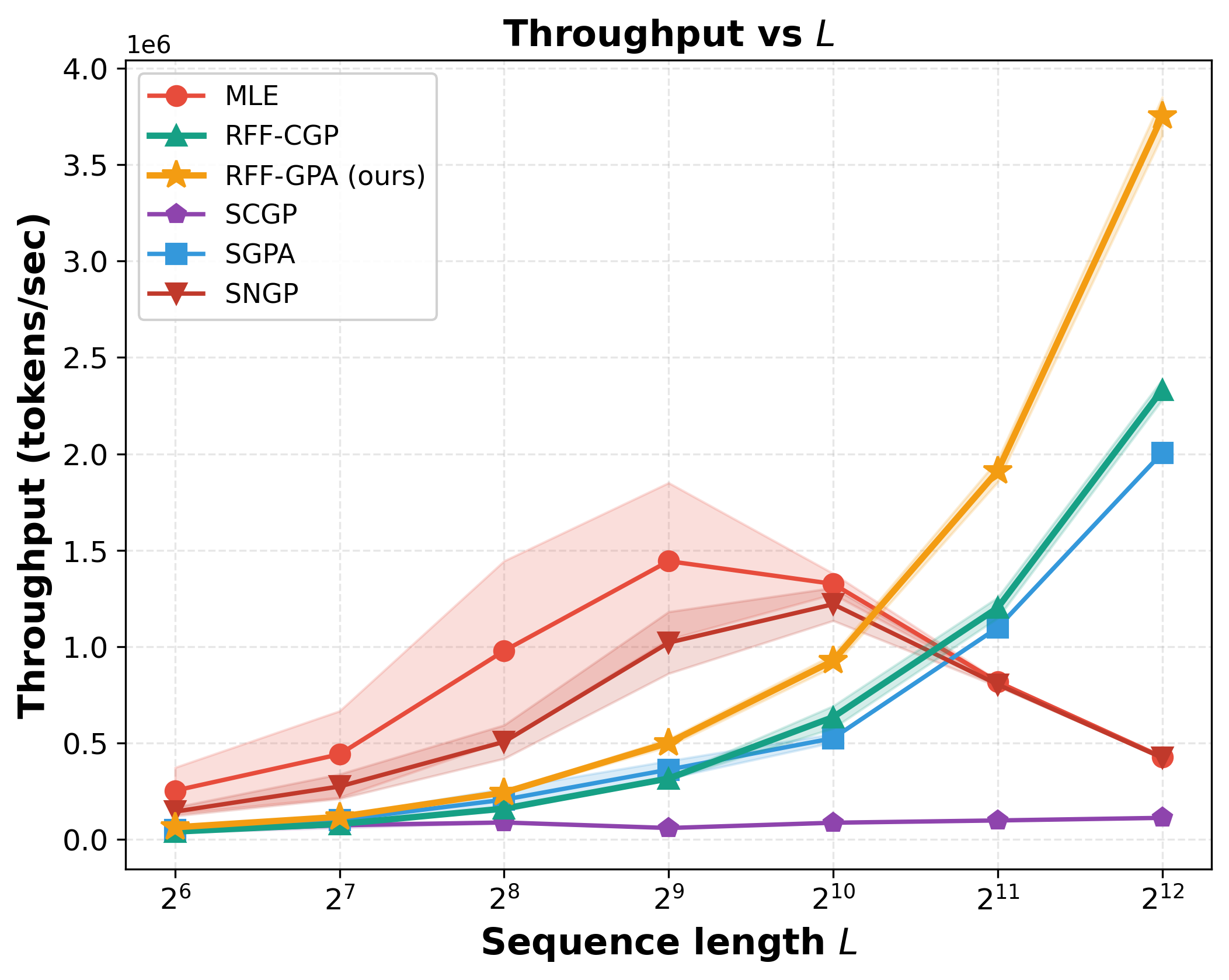}
  \caption{\textbf{Training Time (Left):} Training time per step as a function of sequence length $L$ across all methods (log--log axes; shaded 95\% CI). \textbf{Throughput (Right):} Throughput (tokens/sec) as a function of $L$. Curves for the inducing-point baselines (SGPA, SCGP) are continued past their largest feasible length along their fitted scaling trend.}
  \label{fig:bench2}
\end{figure*}

\paragraph{Calibration.} RFF-GPA delivers strong calibration on the image benchmarks, achieving the best NLL on Fashion-MNIST, CIFAR-10, and SVHN. For ECE, it is best on SVHN and 20~Newsgroups, ties SGPA for the best result on Fashion-MNIST, and is second-best on CIFAR-10 and SST-2. RFF-CGP is particularly strong on the text benchmarks, attaining the best NLL and ECE on Hyperpartisan and SST-2. This modality-dependent pattern suggests that the independent-head RFF-GPA formulation is especially effective for local visual patch interactions, whereas the correlated query--key structure in RFF-CGP can better capture asymmetric semantic interactions in text. Overall, the two random-feature GP variants provide competitive or leading accuracy while yielding well-calibrated predictive uncertainty in a single, linear-time forward pass, without the ensembling or inducing-point machinery required by competing approaches.

\subsection{Efficiency and Scalability}
\label{subsec:efficiency}
We now show that RFF-GPA delivers favorable runtime--calibration behavior compared with inducing-point GP attention. Following the linear-time analysis in Sec.~\ref{sec:rffgpa}, we benchmark a single attention block with a lightweight readout on synthetic inputs while varying the sequence length $L$.\footnote{The full benchmark protocol---backbone, hyperparameters, hardware, and measurement details---is given in Appendix~\ref{app:efficiency}.} Figure~\ref{fig:bench2} reports training time per optimization step and throughput (tokens/sec) as functions of $L$, each averaged over five runs with 95\% confidence intervals.

RFF-GPA and RFF-CGP scale linearly with $L$, thanks to the RFF approximation that only inverts an $M\times M$ matrix. In contrast, the MLE and SNGP softmax heads scale quadratically with sequence length, while inducing-point variants such as SGPA and SCGP exhibit the expected super-linear scaling and incur steep computational costs once $L \gtrsim 500$. These trends are consistent with the complexity analysis in Sec.~\ref{sec:rffgpa}.

The key reason is that the random-feature dimension $M$ is fixed independently of the sequence length, so increasing $L$ only adds feature evaluations and matrix multiplications involving $\Phi_Q$ and $\Phi_K$. The expensive inverse remains in the feature space rather than the token space. By contrast, inducing-point GP attention still materializes sequence-dependent kernel matrices and repeatedly applies kernel operations whose memory footprint grows with $L$, which explains the sharp throughput degradation of SGPA and SCGP at longer sequences. Thus, the empirical scaling curves reflect the intended computational design rather than an implementation artifact. This also makes the cost predictable across batch sizes and sequence lengths, since the dominant operations depend on the chosen feature budget rather than on a growing token--token covariance matrix that must be recomputed for each input.

\begin{table*}[t!]
\centering
\setlength{\tabcolsep}{1.2mm}
\resizebox{\textwidth}{!}{
\begin{tabular}{c||cccccc}
\toprule
METHOD & $s{=}0$ (clean) & $s{=}1$ & $s{=}2$ & $s{=}3$ & $s{=}4$ & $s{=}5$ \\
\midrule
\multirow{1}{*}{METRIC} & \multicolumn{6}{||c}{ACC~$\uparrow$} \\ \midrule\midrule
MLE       & 0.715 $\pm$ 0.008 & 0.681 $\pm$ 0.010 & 0.637 $\pm$ 0.009 & 0.605 $\pm$ 0.009 & 0.564 $\pm$ 0.010 & 0.503 $\pm$ 0.008 \\
SNGP      & 0.727 $\pm$ 0.007 & \underline{0.693 $\pm$ 0.006} & \underline{0.647 $\pm$ 0.001} & \underline{0.613 $\pm$ 0.001} & \underline{0.572 $\pm$ 0.002} & \underline{0.508 $\pm$ 0.003} \\
SGPA      & 0.665 $\pm$ 0.023 & 0.613 $\pm$ 0.023 & 0.556 $\pm$ 0.024 & 0.530 $\pm$ 0.018 & 0.495 $\pm$ 0.014 & 0.441 $\pm$ 0.012 \\
RFF-CGP (ours) & \underline{0.737 $\pm$ 0.003} & 0.664 $\pm$ 0.003 & 0.616 $\pm$ 0.004 & 0.588 $\pm$ 0.004 & 0.548 $\pm$ 0.001 & 0.490 $\pm$ 0.002 \\
\textbf{RFF-GPA (ours)} & \textbf{0.759 $\pm$ 0.001} & \textbf{0.702 $\pm$ 0.001} & \textbf{0.653 $\pm$ 0.002} & \textbf{0.620 $\pm$ 0.002} & \textbf{0.576 $\pm$ 0.003} & \textbf{0.514 $\pm$ 0.002} \\
\midrule
\multirow{1}{*}{METRIC} & \multicolumn{6}{||c}{NLL~$\downarrow$} \\ \midrule\midrule
MLE       & 0.837 $\pm$ 0.022 & 0.946 $\pm$ 0.023 & 1.102 $\pm$ 0.015 & 1.238 $\pm$ 0.010 & \underline{1.421 $\pm$ 0.019} & 1.718 $\pm$ 0.009 \\
SNGP      & 0.807 $\pm$ 0.014 & \underline{0.921 $\pm$ 0.010} & \underline{1.088 $\pm$ 0.023} & \underline{1.234 $\pm$ 0.031} & 1.425 $\pm$ 0.044 & 1.747 $\pm$ 0.060 \\
SGPA      & 0.948 $\pm$ 0.060 & 1.111 $\pm$ 0.062 & 1.298 $\pm$ 0.069 & 1.396 $\pm$ 0.043 & 1.539 $\pm$ 0.020 & 1.770 $\pm$ 0.017 \\
RFF-CGP (ours) & \underline{0.802 $\pm$ 0.005} & 0.990 $\pm$ 0.023 & 1.155 $\pm$ 0.031 & 1.271 $\pm$ 0.031 & 1.445 $\pm$ 0.033 & \underline{1.714 $\pm$ 0.047} \\
\textbf{RFF-GPA (ours)} & \textbf{0.713 $\pm$ 0.007} & \textbf{0.865 $\pm$ 0.012} & \textbf{1.025 $\pm$ 0.016} & \textbf{1.151 $\pm$ 0.023} & \textbf{1.326 $\pm$ 0.028} & \textbf{1.613 $\pm$ 0.027} \\
\midrule
\multirow{1}{*}{METRIC} & \multicolumn{6}{||c}{ECE~$\downarrow$} \\ \midrule\midrule
MLE       & 0.065 $\pm$ 0.005 & 0.079 $\pm$ 0.008 & 0.106 $\pm$ 0.009 & 0.131 $\pm$ 0.008 & 0.164 $\pm$ 0.007 & 0.209 $\pm$ 0.012 \\
SNGP      & 0.063 $\pm$ 0.006 & 0.080 $\pm$ 0.011 & 0.107 $\pm$ 0.016 & 0.132 $\pm$ 0.014 & 0.163 $\pm$ 0.015 & 0.207 $\pm$ 0.013 \\
SGPA      & \textbf{0.025 $\pm$ 0.013} & \underline{0.055 $\pm$ 0.020} & \underline{0.090 $\pm$ 0.023} & \underline{0.104 $\pm$ 0.016} & \underline{0.130 $\pm$ 0.011} & \underline{0.161 $\pm$ 0.009} \\
RFF-CGP (ours) & 0.066 $\pm$ 0.013 & 0.080 $\pm$ 0.020 & 0.108 $\pm$ 0.019 & 0.128 $\pm$ 0.016 & 0.157 $\pm$ 0.014 & 0.197 $\pm$ 0.014 \\
\textbf{RFF-GPA (ours)} & \underline{0.046 $\pm$ 0.003} & \textbf{0.047 $\pm$ 0.005} & \textbf{0.074 $\pm$ 0.005} & \textbf{0.094 $\pm$ 0.006} & \textbf{0.123 $\pm$ 0.004} & \textbf{0.160 $\pm$ 0.003} \\
\bottomrule
\end{tabular}
}
\caption{\textbf{Out-of-distribution robustness on CIFAR-10-C.} Each metric is averaged over the 19 corruption types at severity $s\!\in\!\{1,\dots,5\}$; $s\!=\!0$ is the clean CIFAR-10 test set. Row blocks report ACC~$\uparrow$, NLL~$\downarrow$, and ECE~$\downarrow$. Values are mean~$\pm$~95\% CI over five runs; all models are evaluated \emph{without} re-training. The best result per column is \textbf{bolded} and the second-best is \underline{underlined}.}
\label{table:ood}
\end{table*}

\subsection{Out-of-Distribution Robustness}
\label{subsec:OUT}
We finally evaluate robustness under distribution shift using the CIFAR-10-C benchmark \citep{hendrycks2018benchmarking}, which applies 19 common corruption types (e.g., noise, blur, weather effects) at five severity levels to the CIFAR-10 test set. We take the models trained on clean CIFAR-10 in Sec.~\ref{subsec:ID} and evaluate them on the corrupted test sets \emph{without any re-training or fine-tuning}, so the benchmark measures how gracefully each method degrades once deployed beyond its training distribution.\footnote{We reuse the exact CIFAR-10 checkpoints, backbone, and evaluation protocol from Sec.~\ref{subsec:ID}; the corruption pipeline, normalization, and per-severity averaging are detailed in \appref{app:ood}; baseline and backbone details follow \appref{app:baselines}.} Table~\ref{table:ood} reports the clean CIFAR-10 result at $s\!=\!0$ and, for $s\!\ge\!1$, each metric averaged over the 19 corruptions at every severity level as mean~$\pm$~95\% confidence interval over five runs.

\paragraph{Accuracy.} RFF-GPA is the most accurate method at \emph{every} severity level, from the clean set ($0.759$) down to the most severe corruptions ($0.514$ at $s{=}5$), consistently ahead of the strong SNGP baseline and improving on the inducing-point SGPA by a large margin (e.g.\ $+7.3$ points at $s{=}5$). This shows that the accuracy advantage of the random-feature GP formulation is preserved, not merely retained, under distribution shift.

\paragraph{Calibration.} The central finding is that RFF-GPA is the best-calibrated method across the corruption range in terms of NLL and ECE: it attains the lowest NLL at every severity and the lowest ECE at every non-clean severity ($s\!\ge\!1$), while remaining second only to SGPA on the clean set. Crucially, SGPA reaches competitive ECE only with a substantial loss in accuracy, whereas RFF-GPA combines the best accuracy with the strongest aggregate NLL/ECE. This distinction matters because calibration under shift is meaningful only when the model still preserves useful predictive power: a method can appear well calibrated by becoming uniformly uncertain, but that behavior is less useful if accuracy collapses. RFF-GPA avoids this failure mode by reducing overconfidence while retaining the strongest predictions across severity levels. The deterministic MLE head degrades sharply in NLL as severity grows ($0.84\!\rightarrow\!1.72$), reflecting increasingly overconfident predictions under corruption. Overall, the CIFAR-10-C results confirm that the random-feature GP framework yields reliable uncertainty under distribution shift while preserving predictive performance.

\setlength{\textfloatsep}{8pt plus 2pt minus 2pt}
\begin{figure}[t!]
\centering
\begingroup
\setlength{\tabcolsep}{0pt}
\renewcommand{\arraystretch}{1.0}

\newcommand{\spimg}[1]{\includegraphics[width=0.176\columnwidth]{#1}}

\newcommand{\sphead}[1]{{\fontsize{6.4}{7.0}\selectfont\bfseries #1}}

\newcommand{\splabel}[1]{{\fontsize{6.2}{6.8}\selectfont #1}}

\newcommand{\sprowlabel}[1]{{\fontsize{6.2}{6.8}\selectfont\itshape #1}}

\newcommand{\sptick}[1]{{\fontsize{5.2}{5.8}\selectfont #1}}

\begin{tabular*}{\columnwidth}
{@{\extracolsep{\fill}}ccccc@{}}

\sphead{Clean} &
\sphead{Clean $U$} &
\sphead{Local noise} &
\sphead{Corrupted $U$} &
\sphead{$\Delta U$}
\\

\addlinespace[0.8mm]

\multicolumn{5}{@{}l@{}}{\sprowlabel{Object perturbation}}
\\[-0.2mm]

\shortstack[c]{
  \spimg{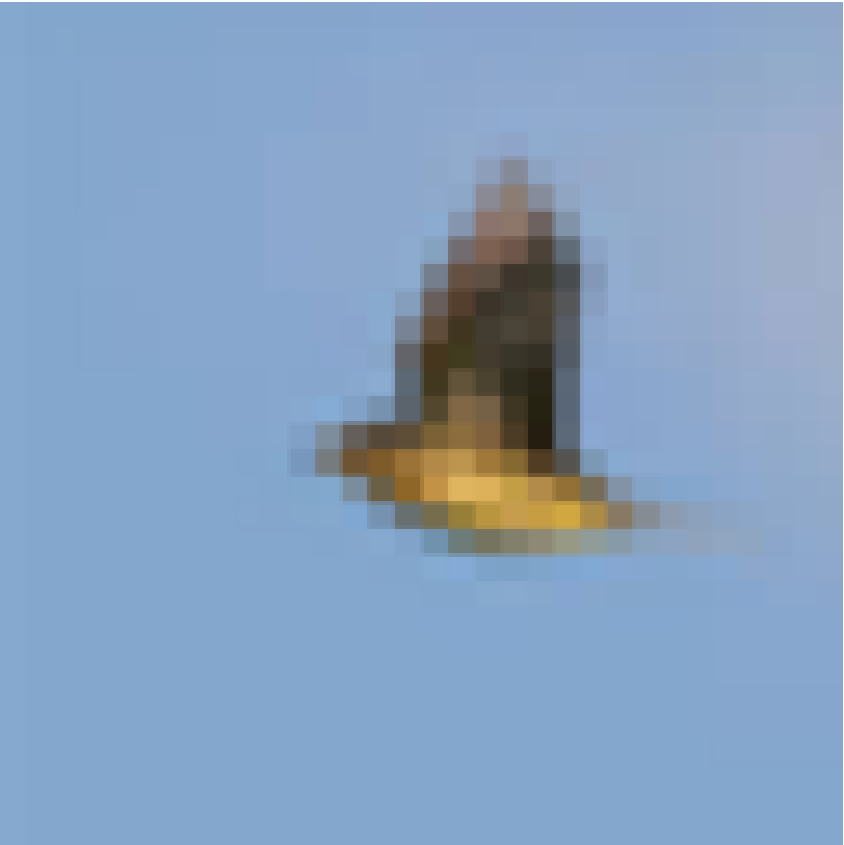}\\[-0.1mm]
  \splabel{\emph{bird}, 85.4\%}
}
&
\shortstack[c]{
  \spimg{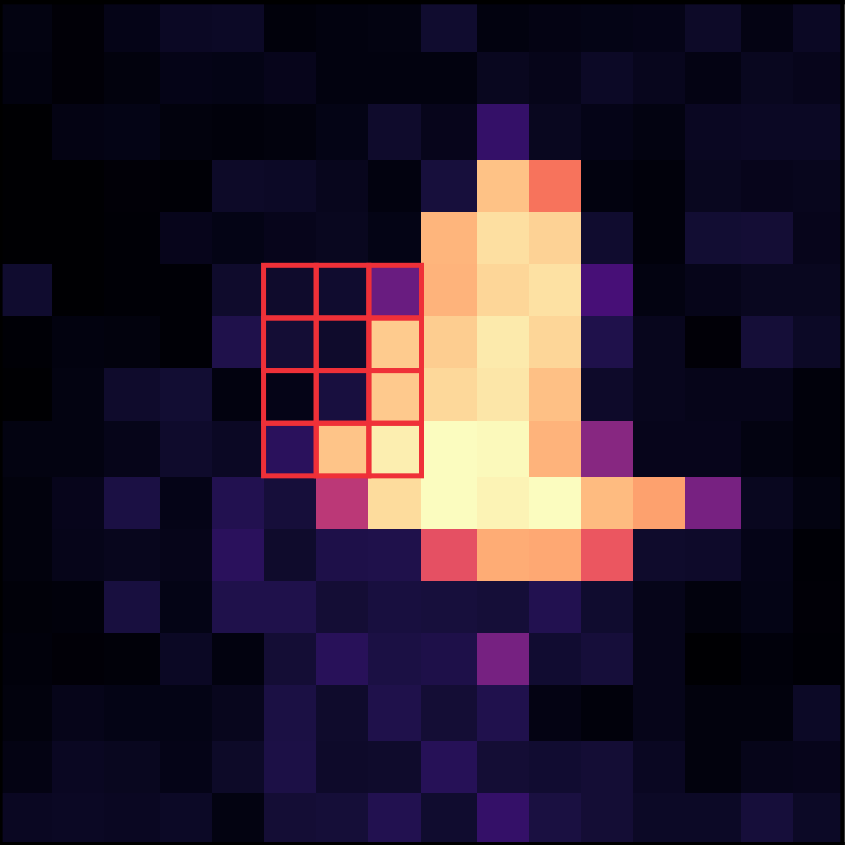}\\[-0.1mm]
  \phantom{\splabel{\emph{bird}, 85.4\%}}
}
&
\shortstack[c]{
  \spimg{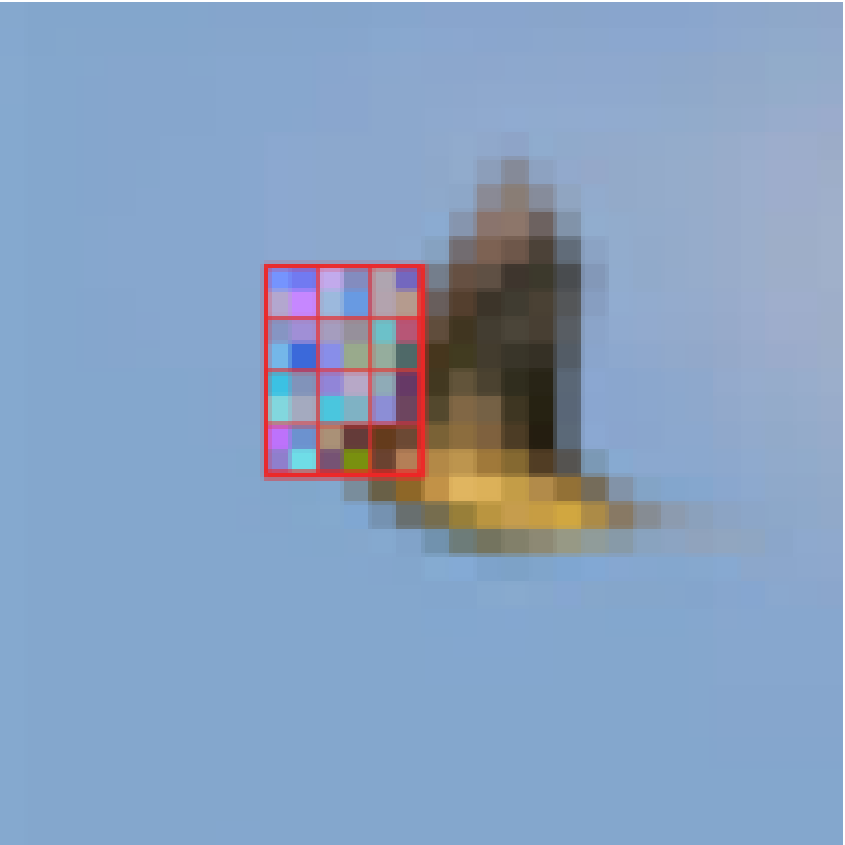}\\[-0.1mm]
  \splabel{\emph{bird}, 66.4\%}
}
&
\shortstack[c]{
  \spimg{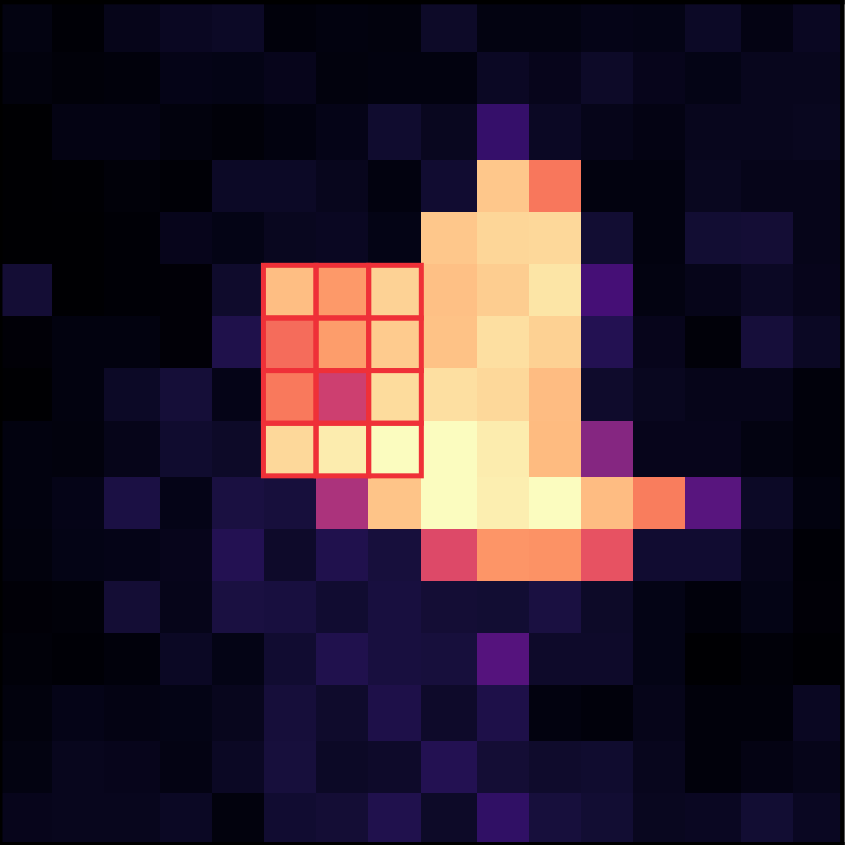}\\[-0.1mm]
  \phantom{\splabel{\emph{bird}, 85.4\%}}
}
&
\shortstack[c]{
  \spimg{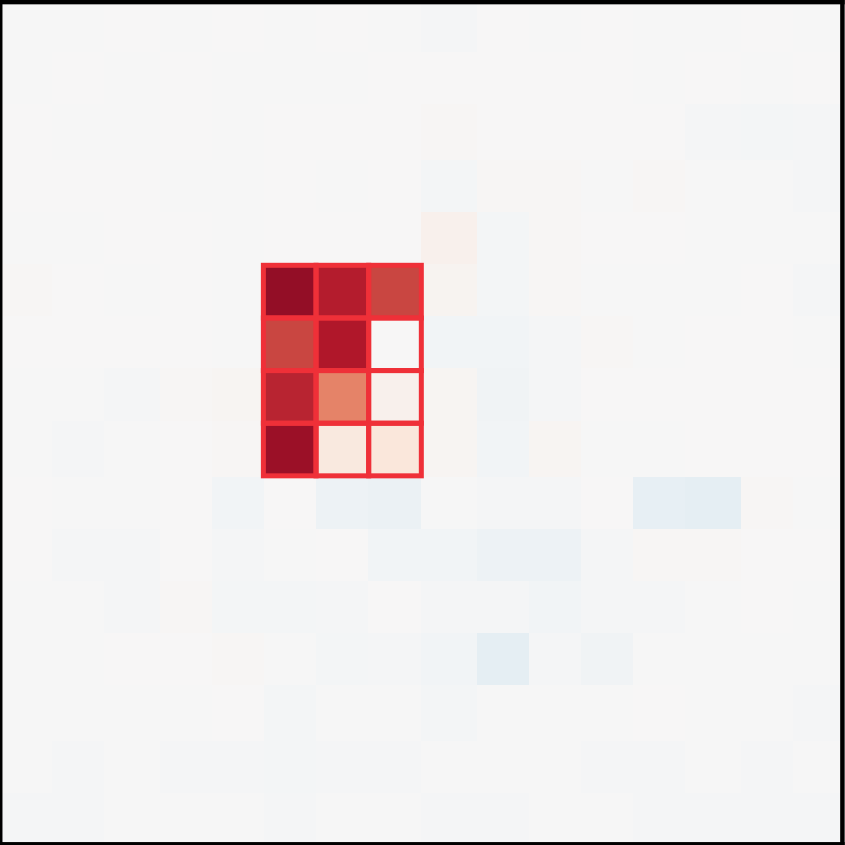}\\[-0.1mm]
  \phantom{\splabel{\emph{bird}, 85.4\%}}
}
\\

\addlinespace[0.7mm]
\cmidrule(lr){1-5}
\addlinespace[0.7mm]

\multicolumn{5}{@{}l@{}}{\sprowlabel{Background perturbation}}
\\[-0.2mm]

\shortstack[c]{
  \spimg{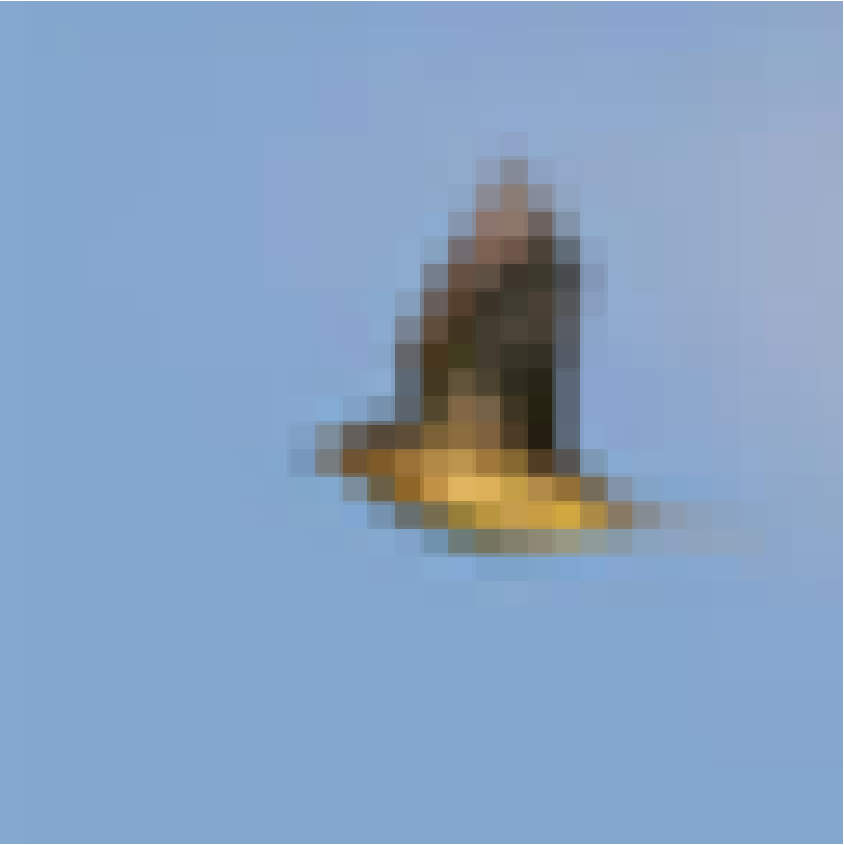}\\[-0.1mm]
  \splabel{\emph{bird}, 85.4\%}
}
&
\shortstack[c]{
  \spimg{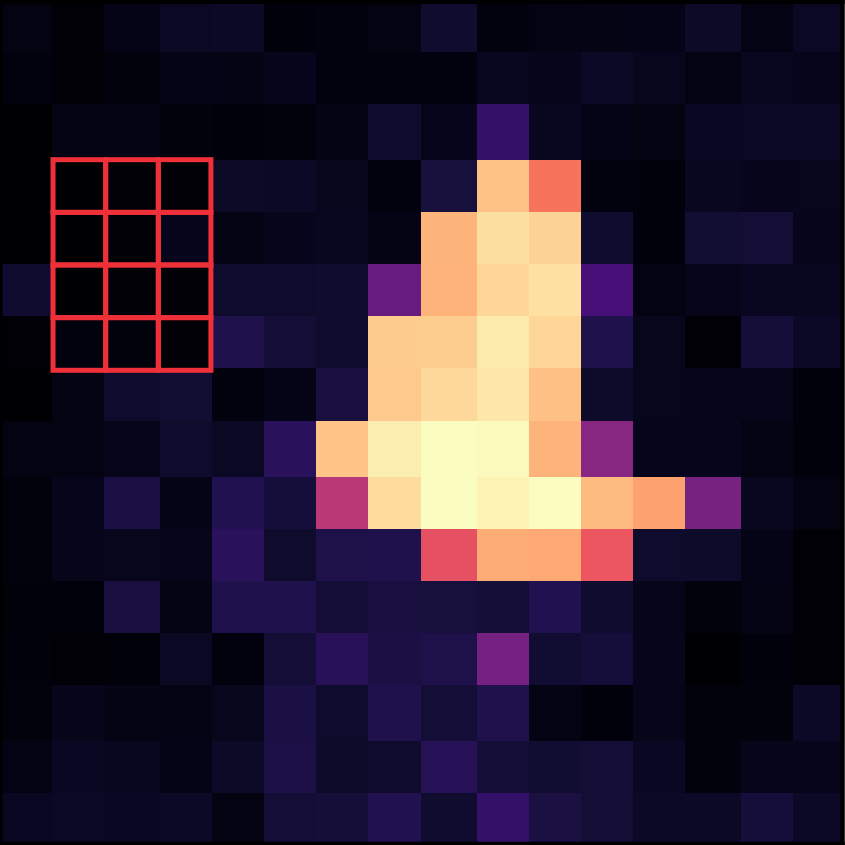}\\[-0.1mm]
  \phantom{\splabel{\emph{bird}, 85.4\%}}
}
&
\shortstack[c]{
  \spimg{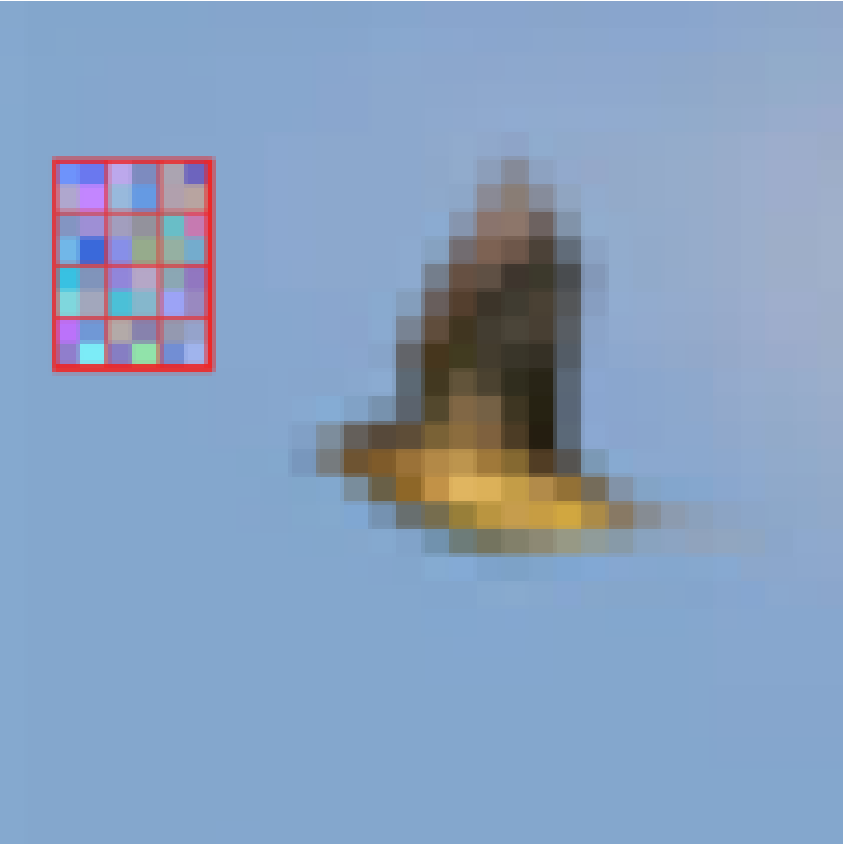}\\[-0.1mm]
  \splabel{\emph{bird}, 75.0\%}
}
&
\shortstack[c]{
  \spimg{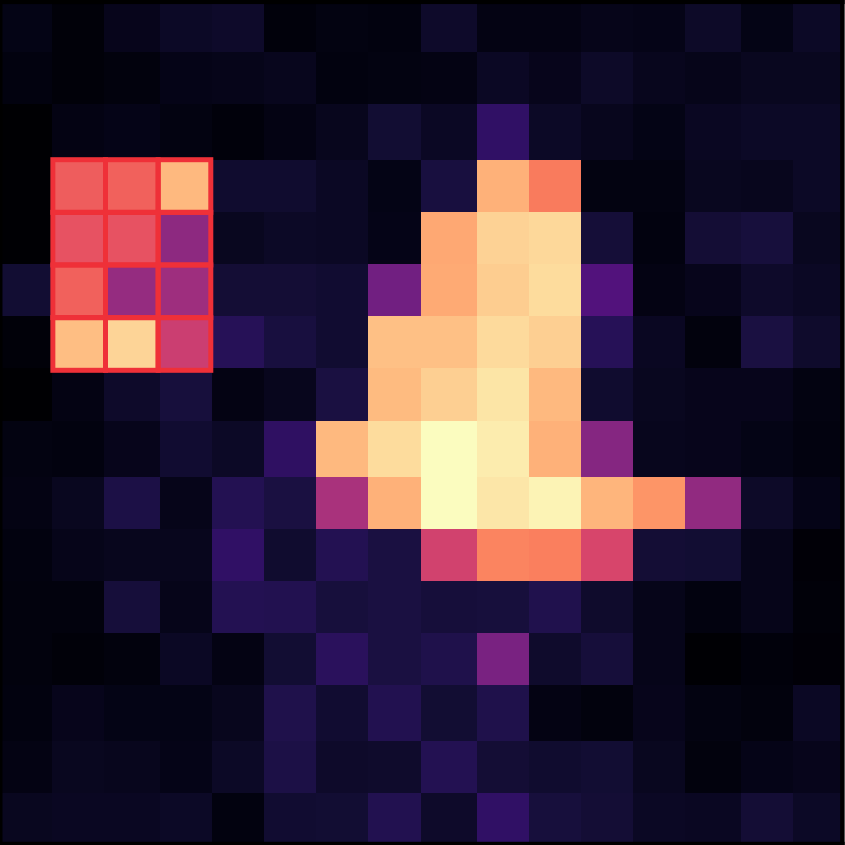}\\[-0.1mm]
  \phantom{\splabel{\emph{bird}, 85.4\%}}
}
&
\shortstack[c]{
  \spimg{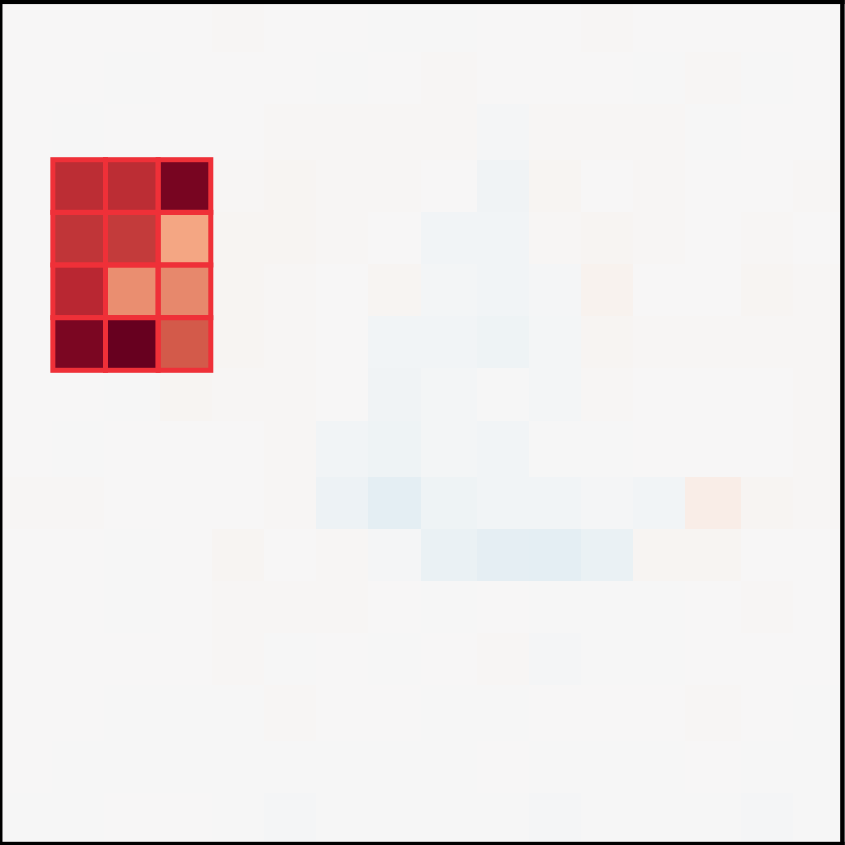}\\[-0.1mm]
  \phantom{\splabel{\emph{bird}, 85.4\%}}
}

\end{tabular*}

\vspace{0.8mm}

\makebox[\columnwidth][c]{\begin{minipage}[t]{0.40\columnwidth}
\centering
{\fontsize{5.8}{6.4}\selectfont
 Posterior variance $U$}\\[-0.2mm]
\includegraphics[
  width=0.92\linewidth,
  height=1.15mm
]{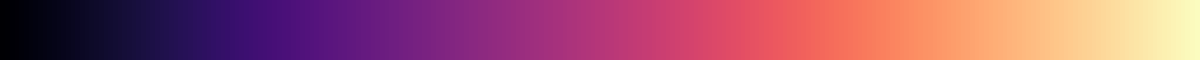}\\[-1.0mm]
\makebox[0.92\linewidth]{\sptick{0.00}\hfill
  \sptick{0.10}\hfill
  \sptick{0.20}}
\end{minipage}
\hspace{0.045\columnwidth}
\begin{minipage}[t]{0.40\columnwidth}
\centering
{\fontsize{5.8}{6.4}\selectfont
 Change in variance $\Delta U$}\\[-0.2mm]
\includegraphics[
  width=0.92\linewidth,
  height=1.15mm
]{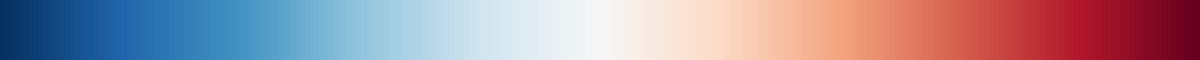}\\[-1.0mm]
\makebox[0.92\linewidth]{\sptick{$-0.15$}\hfill
  \sptick{0}\hfill
  \sptick{0.15}}
\end{minipage}}

\endgroup

\vspace{-0.4mm}

\caption{\textbf{Localized token-wise posterior uncertainty.}
Object perturbation (top) and matched background perturbation
(bottom). Red grids mark the $12$ perturbed tokens; clean and
corrupted uncertainty maps use a shared scale.}
\label{fig:spatial_uncertainty}
\end{figure}

\paragraph{Spatial uncertainty.}
We next examine whether the token-wise GP posterior variance,
$\diag(\bm{\Sigma})$, responds locally to input ambiguity.
Figure~\ref{fig:spatial_uncertainty} shows final-layer uncertainty
before and after adding Gaussian noise to only $12$ of the $256$
CIFAR-10 patch tokens. Variance is averaged over four heads and
$30$ paired Monte Carlo passes, with the same random-feature draw
reused for each clean/corrupted pair. For both an object-side
perturbation and a matched background perturbation, the
prediction remains \emph{bird}, while positive
$\Delta U=U_{\mathrm{corrupt}}-U_{\mathrm{clean}}$ is concentrated
around the perturbed tokens. This provides a spatial complement to
the image-level calibration results above.

\FloatBarrier
\section{Conclusion}
\label{sec:conclusion}
In this paper, we introduced a Gaussian process Transformer where we model the attention kernel using a random Fourier feature approximation. By taking advantage of the low-rank factorization of the attention module, we can obtain a fully Bayesian model that is scalable with respect to the sequence length. We have demonstrated the effectiveness of the RFF-GPA on image and text data sets, especially with respect to uncertainty-aware metrics, compared to the vanilla Transformer and other competing GP-based methods. In future work, we hope to further enrich the approximate kernel structure by modeling non-stationarity across each head, or by assuming a deep generative model as a prior on the random frequencies.

\appendix
\onecolumn
\section*{\centering\textsc{Appendices}}

\section*{\centering\textsc{Appendix Contents}}
\begin{center}
\begin{tabular}{ll}
Appendix~\ref{app:empirical_bayes} & Clarification on the Empirical Bayes Treatment of Spectral Points \\
Appendix~\ref{app:baselines} & Baseline Methods \\
Appendix~\ref{app:datasets} & Datasets \\
Appendix~\ref{app:experimental_design} & Model Configuration and Experimental Design \\
Appendix~\ref{app:efficiency} & Efficiency and Scalability Benchmark \\
Appendix~\ref{app:ood} & Out-of-Distribution Robustness Protocol \\
\end{tabular}
\end{center}

\section{Clarification on the Empirical Bayes Treatment of Spectral Points}
\label{app:empirical_bayes}

In principle, one could introduce a more flexible variational posterior
$q_\phi(\bm\Omega)$ for the spectral points, jointly with the latent Gaussian
attention distribution $p(\Y\mid \X,\bm\Omega)$. The resulting ELBO would take
the general form
\begin{equation}
\mathcal{L}_{\mathrm{ELBO}}
=
\mathbb{E}_{q_\phi(\bm\Omega)}
\mathbb{E}_{p(\Y\mid \X,\bm\Omega)}
\big[\log p(\bm{\mathcal{Y}}\mid \Y)\big]
-
\mathrm{KL}\!\left(q_\phi(\bm\Omega)\,\|\,p(\bm\Omega)\right).
\end{equation}

However, in our setting the spectral points $\bm\Omega$ only influence the
model through the GP prior structure in the attention kernel.
We empirically observed that optimizing the ELBO with a highly expressive
$q_\phi(\bm\Omega)$ tends to drive the KL term toward zero, yielding
$q_\phi(\bm\Omega)\approx p(\bm\Omega)$.
In other words, the variational posterior typically collapses back to the
parametric prior family, while introducing additional variance and instability
during training.

Motivated by these observations, we adopt the simplified assumption
$q(\bm\Omega)=p(\bm\Omega)$ and directly optimize the prior parameters
(e.g., mean and diagonal covariance). This provides a lightweight and stable
empirical Bayes treatment of the spectral density, while retaining the
probabilistic interpretation of the GP-based attention module.

\section{Baseline Methods}
\label{app:baselines}

This appendix summarizes the baseline methods used in our experiments, together with their modeling assumptions and sources of uncertainty.
All baselines are implemented using the same backbone architecture and training protocol unless otherwise stated.

\paragraph{\underline{\textbf{SGPA (Sparse Gaussian Process Attention).}}}
SGPA performs Bayesian inference directly in the output space of multi-head attention by replacing the scaled dot-product similarity with a valid kernel,
and placing a sparse variational Gaussian process (SVGP) posterior over the attention outputs.
Concretely, SGPA models the attention output as a GP mapping with a kernel matrix $K_{QK}$ computed from queries and keys, and uses inducing variables to obtain a scalable variational posterior.
This yields calibrated predictive uncertainty while keeping the computation tractable via sparse GP machinery
\citep{chen2023calibrating}.

\paragraph{SCGP (Sparse Correlated Gaussian Process).}
SCGP is a scalable correlated-GP formulation for kernel attention, derived from correlated Gaussian-process Transformer models.
It introduces a sparse approximation (via inducing inputs) to mitigate the cubic cost of correlated GP inference, while retaining correlation structure in the GP treatment of kernel attention.
In practice, SCGP computes kernel-attention statistics through a correlated GP posterior with a sparse scheme, improving scalability compared to the full correlated GP formulation
\citep{bui2025revisiting}.

\paragraph{\underline{\textbf{SNGP (Spectral-normalized Neural Gaussian Process).}}}
SNGP is a neural Gaussian process model that combines a deterministic deep neural network with a GP-inspired output layer.
Specifically, the penultimate layer features are treated as inputs to a GP classifier,
and spectral normalization is applied to the network weights to ensure distance awareness and improved uncertainty estimates.
Unlike fully Bayesian neural networks, SNGP performs deterministic training while approximating predictive uncertainty through a GP-style output covariance,
making it a strong and scalable baseline for uncertainty estimation in deep models
\citep{liu2020simple}.

\paragraph{\underline{\textbf{MCD (Monte Carlo Dropout).}}}
Monte Carlo dropout (MCD) is a widely used Bayesian approximation for deep neural networks,
where dropout is interpreted as approximate variational inference over network weights.
At test time, multiple stochastic forward passes are performed with dropout enabled,
and predictive uncertainty is estimated from the empirical mean and variance of the outputs:
\begin{equation}
p(y \mid x) \approx \frac{1}{T} \sum_{t=1}^{T} p(y \mid x, \mathbf{W}_t),
\end{equation}
where $\mathbf{W}_t$ denotes a random realization of the network weights induced by dropout.
MCD captures parameter-level (epistemic) uncertainty without modifying the attention mechanism itself,
and serves as a standard baseline for uncertainty-aware Transformers
\citep{gal2016dropout}.

\paragraph{\underline{\textbf{MLE (Deterministic Maximum-Likelihood Head).}}}
The MLE baseline corresponds to a standard deterministic Transformer trained by maximizing the likelihood (or equivalently minimizing cross-entropy loss)
with a softmax output layer.
Predictions are obtained from a single forward pass, and no explicit uncertainty modeling is performed beyond the softmax probabilities.
This baseline serves as a reference point for evaluating the benefits of uncertainty-aware modeling in terms of calibration and robustness.

\paragraph{\underline{\textbf{MLE+Temp (Temperature Scaling).}}}
Temperature scaling is a simple, single-parameter post-hoc calibration method applied to the trained MLE model~\citep{guo2017calibration}.
A scalar temperature $T>0$ is fit on the validation set by minimizing the negative log-likelihood of the temperature-scaled logits $\mathbf{z}/T$,
and predictions at test time use the rescaled softmax $\mathrm{softmax}(\mathbf{z}/T)$.
Since it only rescales confidences without changing the predicted class, MLE+Temp leaves accuracy unchanged while typically improving calibration,
and provides a strong, essentially free reference for the calibration metrics.
Because it is purely post-hoc, temperature scaling is orthogonal to and can be composed with any of the methods considered here, including ours.

\section{Datasets}
\label{app:datasets}

We evaluate on seven publicly available classification benchmarks covering both image and text modalities, summarized in Table~\ref{tab:datasets}. Unless noted otherwise, we hold out a fixed validation split shared across all methods and seeds for model selection and temperature fitting.

\begin{table*}[t]
\centering
\small
\setlength{\tabcolsep}{3pt}
\begin{tabular}{p{0.18\textwidth} p{0.38\textwidth} p{0.22\textwidth} p{0.14\textwidth}}
\toprule
\textbf{Name} & \textbf{Description} & \textbf{Size} & \textbf{Citation/source} \\
\midrule
Fashion-MNIST & Grayscale $28{\times}28$ clothing-image classification with $10$ classes; patch size $4{\times}4$ gives sequence length $49$. & $60$k train / $10$k test & \citet{xiao2017fashion} \\
CIFAR-10 & Color $32{\times}32$ natural-image classification with $10$ classes; patch size $2{\times}2$ gives sequence length $256$. & $50$k train / $10$k test & \citet{krizhevsky2009learning} \\
SVHN & Real-world $32{\times}32$ street-view house-number digit classification with $10$ classes; sequence length $256$. & $\sim$73k train / $26$k test & \citet{netzer2011reading} \\
CIFAR-10-C & Distribution-shift benchmark built from CIFAR-10 with common image corruptions at multiple severities. & $19$ corruptions $\times$ $5$ severities $\times$ $10$k images & \citet{hendrycks2018benchmarking} \\
20~Newsgroups & Topic classification over newsgroup posts with $20$ classes; text is truncated/padded to $512$ tokens. & $\sim$18k documents & \citet{lang1995newsweeder} \\
Hyperpartisan & Binary news-bias classification using the by-article split; text is truncated/padded to $512$ tokens. & Small by-article split & \citet{kiesel2019semeval} \\
SST-2 & Binary sentiment classification from GLUE; we evaluate on the development set because test labels are withheld. & $\sim$67k train; dev evaluation & \citet{socher2013recursive} \\
\bottomrule
\end{tabular}
\caption{Datasets used in the in-distribution and out-of-distribution experiments. Images are patchified inside the Vision-Transformer backbone and standardized per channel; text uses a shared WordPiece tokenizer.}
\label{tab:datasets}
\end{table*}

\section{Model Configuration and Experimental Design}
\label{app:experimental_design}

\paragraph{Overall design.}
Across all image and text benchmarks in Table~\ref{tab:datasets}, we keep the backbone, optimizer, training schedule, validation split, and evaluation metrics fixed within each dataset, changing only the attention or output-layer uncertainty mechanism. This controlled design ensures that differences among MLE, MLE+Temp, MCD, SNGP, SGPA, RFF-GPA, and RFF-CGP reflect the uncertainty mechanism rather than unrelated architectural or tuning choices.

\paragraph{Image classification setup.}
For Fashion-MNIST, CIFAR-10, SVHN, and CIFAR-10-C, we use a compact Vision-Transformer-style encoder with patch embeddings, learned positional embeddings, a learnable [CLS] token, pre-norm residual blocks, and a classification head on the [CLS] representation. Images are converted to tensors and standardized per channel using dataset-specific statistics; we do not use aggressive data augmentation so that calibration comparisons are not confounded by augmentation policy. Patch sizes and resulting sequence lengths follow Table~\ref{tab:datasets}.

\paragraph{Text classification setup.}
For 20~Newsgroups, Hyperpartisan, and SST-2, we use the same Transformer classification template with shared WordPiece tokenization, learned positional embeddings, a [CLS]-style pooled representation, and a linear classification head. Text sequences are truncated or padded to the maximum lengths reported in Table~\ref{tab:datasets}. As with the image experiments, all compared uncertainty mechanisms use the same dataset-specific backbone and training protocol. Unless otherwise stated, the text encoder uses the same model scale as the image encoder: 5 Transformer blocks, hidden size 128, MLP dimension 128, 4 attention heads, and dropout 0.1.

\paragraph{Shared model configuration.}
For every dataset, all compared methods use the same dataset-specific Transformer encoder and differ only in the uncertainty mechanism being evaluated.
Thus, MLE, MLE+Temp, MCD, SNGP, SGPA/SCGP, RFF-GPA, and RFF-CGP are compared under matched depth, hidden size, number of heads, dropout, optimizer, validation split, and evaluation protocol for a given benchmark.
For GP-based attention models, the corresponding image or text backbone is instantiated with the GP attention layer in every Transformer block.
The sparse GP attention baselines (SGPA/SCGP) use the same backbone and classification head but replace random features with inducing-point GP attention.
The deterministic MLE baseline uses the identical architecture with standard scaled dot-product attention (\texttt{kernel\_type = scale\_dot}), while SNGP keeps the attention mechanism deterministic and replaces the output head with a stochastic neural GP layer.

Across the GP-based models, we use one Monte Carlo sample per example during training, add a numerical jitter of $10^{-6}$ in Cholesky/Woodbury solves, and anneal the KL regularization coefficient linearly from 0 to 1 over training via
\[
\lambda_{\text{KL}}(e) = \min\!\bigl(1,\ 2e / E\bigr),
\]
where $e$ is the epoch index and $E$ is the total number of training epochs for that dataset.
Randomness is controlled via a single seed parameter (\texttt{--seed}), and we train each model over five seeds to compute the reported error bars.

\paragraph{Training protocol.}
All in-distribution experiments use the Adam optimizer
\[
\text{Adam}(\beta_1 = 0.9, \ \beta_2 = 0.999),
\]
with no weight decay, i.e., $\lambda = 0$.
The learning rate is recomputed at the start of each epoch using a piecewise-linear warmup and decay schedule:
we set the initial learning rate to $\eta_{\text{ini}} = 10^{-5}$, the base learning rate to $\eta_{\text{base}} = 5 \times 10^{-4}$, and the minimum learning rate to $\eta_{\min} = 10^{-5}$.
The warmup phase spans $E_{\text{warm}} = 5$ epochs, during which the learning rate increases linearly from $\eta_{\text{ini}}$ to $\eta_{\text{base}}$.
This is followed by a linear decay phase of $E_{\text{decay}} = 480$ epochs down to $\eta_{\min}$.
For image benchmarks, we use a global batch size of $B_{\text{img}} = 128$ for training and $B_{\text{img,test}} = 128$ for validation and test.
For text benchmarks, each method uses the same tokenizer, maximum length, batch size, optimizer, and model-selection split within that dataset.

\paragraph{Image-model instantiation.}
The image encoder used for Fashion-MNIST, CIFAR-10, SVHN, and CIFAR-10-C is a compact Vision Transformer with $L_{\text{img}} = 5$ transformer blocks, hidden size $d_{\text{model}}^{\text{img}} = 128$, MLP dimension $d_{\text{ff}}^{\text{img}} = 128$, and $H_{\text{img}} = 4$ attention heads per layer.
We use pre-norm residual blocks and apply dropout with rate $p_{\text{drop}} = 0.1$ in the attention and MLP layers.
Images are split into non-overlapping patches according to Table~\ref{tab:datasets}, prepended with a learnable [CLS] token, and combined with learned positional embeddings.
For CIFAR-10 and SVHN, $2 \times 2$ patches yield $16 \times 16 = 256$ patch tokens; for Fashion-MNIST, $4 \times 4$ patches yield $7 \times 7 = 49$ patch tokens.

\paragraph{Random-feature and inducing-point settings.}
Across both image and text benchmarks, we use \texttt{keys\_len = 32} for our random-feature GP attention, which yields an effective random-feature dimension $M_{\text{eff}} = 64$ per head via cosine--sine pairs.
RFF-GPA draws $32$ frequencies and forms cosine/sine features, while RFF-CGP uses cosine--sine features from $32$ shared frequencies.
The inducing-point baselines (SGPA/SCGP) use $16$ inducing points under the same backbone and training protocol.

\paragraph{CIFAR-10-C evaluation.}
CIFAR-10 consists of 50{,}000 training and 10{,}000 test images of size $32 \times 32$ across 10 classes.
From the 50{,}000 training images, we hold out 5{,}000 examples as a validation set and train on the remaining 45{,}000 examples.
CIFAR-10-C is used only for inference-time distribution-shift evaluation: we reuse the corresponding CIFAR-10 checkpoints and evaluate on all 19 corruption types at 5 severity levels, reporting averages over corruptions and severities.

\section{Efficiency and Scalability Benchmark}
\label{app:efficiency}

This appendix details the efficiency and scalability protocol summarized in Sec.~\ref{subsec:efficiency}. All measurements probe asymptotic behavior and hardware efficiency independent of data loading, so we use synthetic inputs and benchmark a single attention block rather than an end-to-end model.

\paragraph{Benchmark unit.}
For each method we build one attention layer followed by a lightweight readout (\texttt{LayerNorm} $\rightarrow$ \texttt{GELU} $\rightarrow$ \texttt{Linear} to a scalar), and feed it a synthetic input tensor $X \in \mathbb{R}^{B \times L \times d}$ with $X \sim \mathcal{N}(0, I)$. The attention layers are the exact same implementations used in our in-distribution experiments (Appendix~\ref{app:baselines}); only the surrounding backbone is replaced by the synthetic-input readout so that the measured cost reflects the attention mechanism itself. We sweep the sequence length $L \in \{64, 128, 256, 512, 1024, 2048, 4096\}$ and, for the inducing-point baselines, draw a fresh set of inducing keys per step.

\paragraph{Model configuration.}
To match the in-distribution setup (Appendix~\ref{app:baselines}), all methods use hidden size $d_{\text{model}} = 128$, $H = 4$ attention heads, batch size $B = 16$, and \texttt{float32}. Random-feature methods use an effective feature count of $M = 32$ per head (RFF-GPA: \texttt{feature\_mode=rff} with $\texttt{keys\_len}=16$; RFF-CGP: $M = \texttt{keys\_len}$); inducing-point methods (SGPA, SCGP) use $16$ inducing points; and we set the GP jitter to $10^{-6}$ and dropout to $0.1$. GP variants draw a single Monte-Carlo sample per forward pass at training time and $5$ samples at inference time, while the deterministic MLE and SNGP heads use a single pass.

\paragraph{Timing and memory measurement.}
Each configuration is run with $10$ untimed warmup steps followed by $60$ timed steps; every step performs a full forward pass, backward pass, and \texttt{AdamW} update (learning rate $3\times 10^{-4}$). We use \texttt{torch.cuda.synchronize} around each phase, report the mean per-step time (with forward/backward/optimizer components measured separately), compute throughput as $B\!\cdot\!L$ divided by the mean step time, and read peak memory from \texttt{torch.cuda.max\_memory\_allocated}. For the runtime--calibration comparison in Sec.~\ref{subsec:efficiency}, runtime is reported as time per $1000$ tokens at $L = 1024$, and the corresponding ECE values are taken from the in-distribution CIFAR-10 evaluation in Sec.~\ref{subsec:ID}. All runs are repeated over five seeds.

\paragraph{Hardware and large-$L$ handling.}
All timing and memory results are measured on a single RTX 4090-class GPU (24\,GB). The inducing-point baselines materialize an $L\times L$ kernel and therefore grow quadratically in memory: SCGP exhausts the 24\,GB budget beyond $L = 512$, and SGPA beyond $L = 1024$. For these methods we measure up to their largest feasible length and fit a power-law trend to the measured values ($\text{metric} \propto L^{p}$), consistent with their known $\mathcal{O}(L^{2})$ scaling; the random-feature methods (RFF-GPA, RFF-CGP) and the softmax heads (MLE, SNGP) are measured directly across the entire range.

\section{Out-of-Distribution Robustness Protocol}
\label{app:ood}

This appendix details the CIFAR-10-C evaluation summarized in Sec.~\ref{subsec:OUT} and reported in Table~\ref{table:ood}. The evaluation is \emph{inference-only}: we reuse the exact CIFAR-10 checkpoints, backbone, and metrics from the in-distribution experiment (Sec.~\ref{subsec:ID}, Appendix~\ref{app:baselines}) and never re-train or fine-tune on corrupted data, so the results measure how the trained models generalize to a shifted input distribution rather than the effect of any additional training.

\paragraph{Corruption benchmark.}
We use the standard CIFAR-10-C benchmark \citep{hendrycks2018benchmarking}, which perturbs the $10{,}000$-image CIFAR-10 test set with $19$ corruption types spanning four families---noise (Gaussian, shot, impulse, speckle), blur (defocus, glass, motion, zoom, Gaussian), weather (snow, frost, fog, brightness, spatter), and digital (contrast, elastic transform, pixelate, JPEG compression, saturate)---each applied at five severity levels $s\in\{1,\dots,5\}$. Every corruption/severity pair thus yields a $10{,}000$-image corrupted test set with the original CIFAR-10 labels. We treat the clean CIFAR-10 test set as severity $s=0$, taken directly from the in-distribution evaluation in Sec.~\ref{subsec:ID}. In total each model is evaluated on $19\times5=95$ corrupted test sets plus the clean set.

\paragraph{Preprocessing.}
Corrupted images are provided as raw $32\times32$ RGB arrays. We apply the \emph{same} preprocessing as during training---conversion to $[0,1]$ tensors followed by per-channel standardization with the CIFAR-10 training mean $(0.4914,0.4822,0.4465)$ and standard deviation $(0.2470,0.2435,0.2616)$---and apply no test-time augmentation. Using the training-time normalization statistics is essential: any mismatch would itself introduce a distribution shift and confound the corruption effect we intend to measure. Patchification into a length-$256$ token sequence is handled inside the ViT backbone, identically to the in-distribution setting.

\paragraph{Inference and metrics.}
Each checkpoint is evaluated with the same predictive protocol as in Sec.~\ref{subsec:ID}: GP-based variants (SGPA, RFF-CGP, RFF-GPA) average $10$ Monte-Carlo samples per input, whereas the deterministic MLE and SNGP heads use a single forward pass. For every corruption/severity pair we compute accuracy (ACC), negative log-likelihood (NLL), and expected calibration error (ECE, $15$ equal-width confidence bins) on the predicted class probabilities, using the identical metric implementations as the in-distribution experiment.

\paragraph{Aggregation.}
For each model and severity level we first average every metric over the $19$ corruption types, yielding one per-severity value per run. We repeat the whole evaluation for the five training seeds and report, in Table~\ref{table:ood}, the mean over seeds together with a $95\%$ confidence interval (computed as $1.96\,\hat{\sigma}/\sqrt{5}$). The clean column ($s=0$) is the seed-averaged in-distribution CIFAR-10 result under the same aggregation. This corruption-averaged, per-severity summary follows standard practice for CIFAR-10-C and isolates the effect of increasing corruption strength from the idiosyncrasies of individual corruption types.

\end{document}